\documentclass[10pt,twocolumn,letterpaper]{article}

\usepackage{cvpr}              

\usepackage[T1]{fontenc}
\usepackage{xcolor}
\usepackage{pgf}
\usepackage{listings}

\newcommand{\GoodValue}{0.80}
\newcommand{\MidValue}{0.675} 
\newcommand{\BadValue}{0.55}

\definecolor{GoodColor}{RGB}{102, 194, 165}   
\definecolor{MidColor}{RGB}{252, 229, 155}    
\definecolor{BadColor}{RGB}{239, 138, 98}     

\newcommand{\C}[1]{%
    \pgfmathsetmacro{\BlendFactor}{(#1-\BadValue)/(\GoodValue-\BadValue)}%
    \pgfmathsetmacro{\BlendFactor}{min(max(\BlendFactor,0),1)}%
    \ifdim #1 pt < \MidValue pt
        \pgfmathsetmacro{\LocalBlend}{\BlendFactor / ((\MidValue-\BadValue)/(\GoodValue-\BadValue)) * 100}%
        \colorbox{MidColor!\LocalBlend!BadColor}{#1}%
    \else
        \pgfmathsetmacro{\LocalBlend}{(\BlendFactor - (\MidValue-\BadValue)/(\GoodValue-\BadValue)) / (1 - (\MidValue-\BadValue)/(\GoodValue-\BadValue)) * 100}%
        \colorbox{GoodColor!\LocalBlend!MidColor}{#1}%
    \fi
}

\definecolor{cvprblue}{rgb}{0.21,0.49,0.74}
\usepackage[pagebackref,breaklinks,colorlinks,allcolors=cvprblue]{hyperref}

\def\paperID{4146} 
\def\confName{CVPR}
\def\confYear{2025}

\title{A Comprehensive Study of Decoder-Only LLMs for Text-to-Image Generation}

\author{
    \begin{tabular}{c}
        Andrew Z. Wang\textsuperscript{1,3}\thanks{This work was done during an internship at NVIDIA.} \and
        Songwei Ge\textsuperscript{2}\footnotemark[2] \and
        Tero Karras\textsuperscript{3} \and
        Ming-Yu Liu\textsuperscript{3} \and
        Yogesh Balaji\textsuperscript{3} \and
    \end{tabular} \\
    \begin{tabular}{c}
        \textsuperscript{1}University of Washington \and
        \textsuperscript{2}University of Maryland \and
        \textsuperscript{3}NVIDIA
    \end{tabular}
}

\newlength{\oldtabcolsep}
\begin{document}
\maketitle

\begin{abstract}
Both text-to-image generation and large language models (LLMs) have made significant advancements. However, many text-to-image models still employ the somewhat outdated T5 and CLIP as their text encoders.
In this work, we investigate the effectiveness of using modern decoder-only LLMs as text encoders for text-to-image diffusion models. We build a standardized training and evaluation pipeline that allows us to isolate and evaluate the effect of different text embeddings. We train a total of 27 text-to-image models with 12 different text encoders to analyze the critical aspects of LLMs that could impact text-to-image generation, including the approaches to extract embeddings, different LLMs variants, and model sizes.
Our experiments reveal that the \textit{de facto} way of using last-layer embeddings as conditioning leads to inferior performance.
Instead, we explore embeddings from various layers and find that using
layer-normalized averaging across all layers significantly improves alignment with complex prompts. Most LLMs with this conditioning outperform the baseline T5 model, showing enhanced performance in advanced visio-linguistic reasoning skills. 
\end{abstract}

\section{Introduction}
\label{sec:intro}

The field of text-to-image generation has seen tremendous progress in recent years, driven by improvements in diffusion architectures \cite{esser_scaling_2024, chen_pixart-_2023, liu_playground_2024}, scalable models \cite{podell_sdxl_2023, Peebles2022DiT}, and better training procedures \cite{liu_playground_2024, Karras2022edm, Karras2024edm2}. Central to the success of these models is the use of pre-trained text encoders that transfer natural language prompts into representations suitable for guiding image generation. However, the impact of different text encoder models on the image generation quality remains largely unexplored.

\begin{figure}[t]
\includegraphics[scale=0.34]{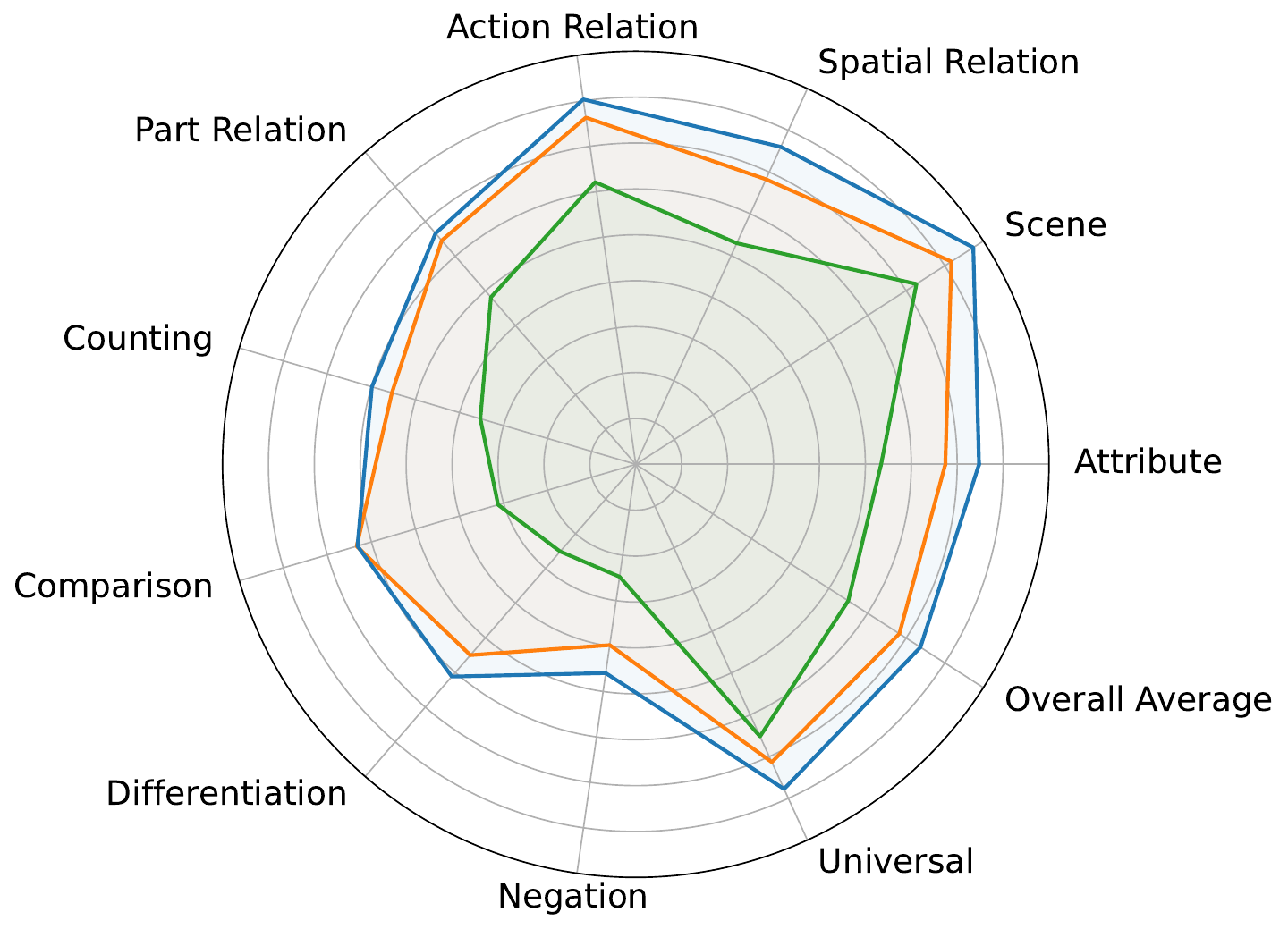}
\centering
\caption{VQA scores for text-to-image models using T5-XXL and Mistral-7B as the text encoders. The models use embeddings extracted from the last layer of T5 (\textit{orange}) and Mistral (\textit{green}), and layer-normalized average embeddings of Mistral (\textit{blue}). Our results show that Mistral performs worse than T5 when only using the last layer. However, using all layers significantly improves Mistral’s performance, surpassing T5 in all aspects.}
\label{fig:radar}
\end{figure}

Most contemporary image generators use traditional text encoders such as T5 \cite{raffel_exploring_2023} or CLIP \cite{radford_learning_2021} for obtaining text embeddings~\cite{podell_sdxl_2023, saharia_photorealistic_2022, balaji_ediff-i_2023, DALLE2, nvidia2024edifyimage}. However, the development of encoder-decoder models has slowed down while there is an increasing interest in decoder-only large language models (LLMs) due to their great scalability. This raises a pressing question of whether these decoder-only LLMs are suitable for text-to-image generation tasks.


In this work, we examine how the rich linguistic representations from decoder-only LLMs can be leveraged to enhance text-to-image generation and investigate the most effective strategies for incorporating them. Our focus is on autoregressive language models that use causal attention and are trained with a next-token prediction objective. We address three primary research questions:
\begin{enumerate}[topsep=2mm, itemsep=2mm]
    \item To what extent can decoder-only LLMs be utilized to enhance text-to-image generation? What methodologies yield the largest improvements?
    \item LLM fine-tuned embedding models have surpassed encoder-decoder models in contextual semantic comprehension across various tasks \cite{muennighoff_mteb_2023}. Can these embedding models contribute to improving text-to-image alignment?
    \item Does increasing the model size of decoder-only LLMs lead to measurable improvements in the text-to-image generation performance?
\end{enumerate}

To address these questions, we develop a standardized text-to-image training and evaluation pipeline based on the Stable Diffusion v2 (SD2) \cite{Rombach_2022_CVPR} architecture. In each experiment, only the text encoder is replaced, while the rest of the architecture and training recipe remains unchanged. This allows us to isolate and understand the effect of different text representations. We explore a wide variety of models as text encoders, including two traditional text-to-image text encoders, seven open-source LLMs, and three fine-tuned embedding models. 
We quantitatively evaluate the models using VQAScore \cite{li_genai-bench_2024} on GenAI-Bench \cite{li_genai-bench_2024}, and conduct in-depth analysis to understand the strengths and limitations of using LLMs as the text encoders. 
%
%
Our main findings are as follows:
\begin{itemize}[topsep=2mm, itemsep=2mm]
    \item \textbf{Using text embeddings from the last layer of the LLM is sub-optimal}:
    To our knowledge, all current text-to-image diffusion models utilize the embeddings from the final layer of the text encoders as the conditional embedding~\cite{saharia_photorealistic_2022, balaji_ediff-i_2023}. However, our results reveal that this approach does not translate effectively to LLMs, leading to inferior results compared to using T5.
    \item \textbf{Aggregating features from multiple layers outperforms using a single layer}: 
    We find that using embeddings normalized and averaged across all layers yields far better performance than relying on any single layer alone (Figure \ref{fig:radar}). This is because each layer within an LLM captures different aspects of linguistic information, so using averaged embeddings can combine the strengths of every layer to create a richer and more comprehensive representation \cite{jawahar_what_2019, dar2023analyzingtransformersembeddingspace, liu_fantastic_2024}.
    \item \textbf{LLM-based embedding models sometimes outperform the base models}: Our results suggest that fine-tuned embedding models hold the potential for improving text-to-image generation. However, such a performance improvement is not always observed across all the embedding and base models. 
    Despite these mixed outcomes, strong results are achieved with bge-Gemma2 \cite{chen_bge_2024} using layer-normalized average embeddings, our best-performing model, highlighting promise in leveraging embedding models for text-to-image generation.
    \item \textbf{Scaling up the LLM is beneficial, but not across all aspects}: Increasing the model size of LLMs consistently leads to improved performance. However, we observe that model size does not uniformly enhance all aspects of compositional text-to-image generation. These results suggest that simply scaling model size may not be the most efficient approach for improving performance across all skills, highlighting the potential of alternative strategies, such as hybrid models or skill-specific fine-tuning.
\end{itemize}
\section{Related work}
\label{sec:related}

\setlength{\tabcolsep}{1pt}
\renewcommand{\arraystretch}{0.5}
\begin{figure*}[t]
    \centering
    \begin{tabular}{cccc}
        {\centering CLIP (\emph{last layer})} & {\centering T5-XXL (\emph{last layer})} & {\centering Mistral (\emph{norm avg})} & {\centering bge-Gemma2 (\emph{norm avg})} \\[2mm] 
        \includegraphics[width=0.245\textwidth]{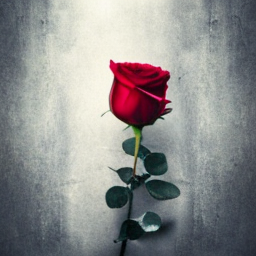} &
        \includegraphics[width=0.245\textwidth]{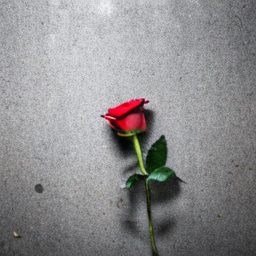} &
        \includegraphics[width=0.245\textwidth]{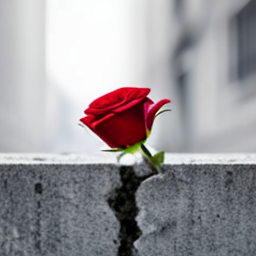} &
        \includegraphics[width=0.245\textwidth]{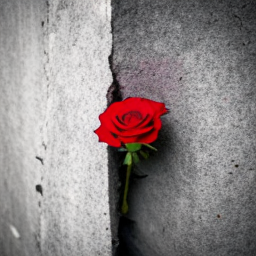} \\[2mm]
        \multicolumn{4}{c}{\emph{A single, vibrant red rose blooms defiantly through a \textbf{narrow crack in the weathered, grey concrete}.}} \\[1mm] 
        \multicolumn{4}{c}{\emph{Its velvety petals unfurl gracefully, reaching for the sunlight that filters weakly through the urban haze.}} \\[2mm] 
        \includegraphics[width=0.245\textwidth]{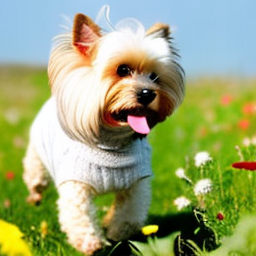} &
        \includegraphics[width=0.245\textwidth]{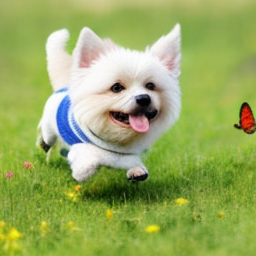} &
        \includegraphics[width=0.245\textwidth]{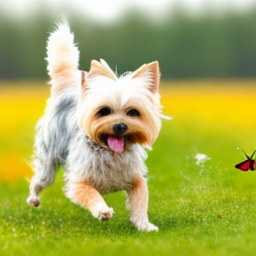} &
        \includegraphics[width=0.245\textwidth]{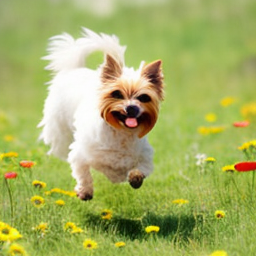} \\[2mm]
        \multicolumn{4}{c}{\emph{A small dog \textbf{not in} a tiny sweater, playing joyfully \textbf{without any clothes}.  The fluffy white dog, with big brown eyes and}} \\[1mm]
        \multicolumn{4}{c}{\emph{floppy ears, bounds through a sun-drenched field of wildflowers. Its tongue lolls out in pure happiness as it chases a}} \\[1mm]
        \multicolumn{4}{c}{\emph{bright red butterfly, its tiny paws barely touching the ground.}} \\[2mm]
\end{tabular}
\caption{Visual comparison of images generated with different text encoders. We use last-layer embeddings (\textit{last layer}) from the text encoders of \href{https://huggingface.co/laion/CLIP-ViT-H-14-laion2B-s32B-b79K}{CLIP-ViT-H/14} (354M) and \href{https://huggingface.co/google-t5/t5-11b}{T5-XXL} (4.7B). We also use average layer-normalized embeddings (\textit{norm avg}) from the pre-trained LLM \href{https://huggingface.co/mistralai/Mistral-7B-v0.1}{Mistral-7B} (7B) and the fine-tuned embedding model bge-Gemma2 (\href{https://huggingface.co/BAAI/bge-multilingual-gemma2}{bge-multilingual-gemma2}; 9B). Mistral and bge-Gemma2 can handle complex reasoning tasks such as negation (panel below) compared to models like CLIP and T5.}
\label{fig:visual_comparison}
\end{figure*}
\setlength{\tabcolsep}{\oldtabcolsep}
\renewcommand{\arraystretch}{1}

\paragraph{Text encoders in text-to-image generation.} In recent years, a variety of text encoders have been explored for training large-scale diffusion models. DALLE2~\cite{DALLE2} and Stable Diffusion~\cite{Rombach_2022_CVPR} use text embeddings from the CLIP~\cite{radford_learning_2021} model that is trained using an image-text alignment objective. Imagen~\cite{saharia_photorealistic_2022} showed that the embeddings from pure language models such as T5~\cite{raffel_exploring_2023} can be used for image generation and that scaling the language models can improve the image generation performance. eDiff-I~\cite{balaji_ediff-i_2023} observed that the use of CLIP and T5 can give complementary benefits and propose the use of a combination of both these embeddings. Liu~et~al.~\cite{liu2022character} showed that using character-aware tokenizers like ByT5~\cite{xue2022byt5} can improve text rendering while generating images.


\paragraph{Decoder-only LLMs.} Several studies have looked into exploiting LLMs for text-to-image generation. Earlier works utilized LLMs as inputs through adapter connectors \cite{hu_ella_2024, he2024marsmixtureautoregressivemodels} and explored their sequence-to-sequence capabilities to revise prompts \cite{chen_textdiffuser-2_2023} or build scene layouts \cite{lian2024llmgroundeddiffusionenhancingprompt}. The recent concurrent work of Playground-v3 \cite{liu_playground_2024} conditions different blocks of the diffusion transformer model using embeddings from intermediate layers of a Llama3 \cite{dubey_llama_2024} model. Lumina-T2X~\cite{gao2024lumina} uses the output layer of a LLama-7B~\cite{touvron2023llama} model as the text encoder, while Lumina-Next~\cite{zhuo2024lumina} and Sana~\cite{xie2024sana} use the embeddings from Gemma-2-2B~\cite{team_gemma_2024} model.

\paragraph{LLM interpretability.} As we are interested in analyzing the best ways to utilize LLMs, a relevant line of research is on analyzing the linguistic representations captured across different layers within LLMs \cite{jawahar_what_2019, tenney_what_2019, dar2023analyzingtransformersembeddingspace, liu_fantastic_2024, zhang_investigating_2024}. Some studies have addressed the contextual limitations imposed by causal attention masks in autoregressive models \cite{springer_repetition_2024}, while others have examined approaches to best extract embedding representations from LLMs \cite{li_towards_2023, wang2024textembeddingsweaklysupervisedcontrastive, behnamghader_llm2vec_2024, wang_improving_2024, chen_bge_2024, jin2024analyzingrolesemanticrepresentations}.

While these recent works investigate the use of various text encoders for image generation, their models are trained on different datasets (often proprietary) and with varying training setups, making it difficult to isolate the role of text encoders. The goal of this work is to perform a systematic study on the effect of different text encoders, aiming to deepen understanding of how text representations influence text-image alignment and generation quality.
\section{Experimental setup}
\label{sec:setup}

\subsection{Training pipeline}

To understand the impact of text encoders in text-to-image diffusion models, we first establish a standard model training pipeline, enabling us to isolate the effect of text encoders from other factors, like diffusion model architecture, training data, and compute.
For all our experiments, we utilize the U-Net based latent diffusion architecture from Stable Diffusion \cite{Rombach_2022_CVPR}. We freeze the autoencoder and only train the U-Net model~\cite{ronneberger_u-net_2015}, in which text embeddings are conditioned  using cross attention blocks. To accommodate the different text embedding dimension sizes from various models, we introduce a linear projection layer with $1024$ output features before cross-attention in the U-Net. In each experiment, we replace the text encoder with a different LLM or fine-tuned embedding model.
The exact training recipe, including batch size, diffusion process, and other hyperparameters are also adapted from Stable Diffusion \cite{Rombach_2022_CVPR}. Please find more details in the supplementary material.

We utilize a $46$ million text-image pair subset of the LAION-Aesthetics dataset~\cite{schuhmann_laion-5b_2022} as the training data. For enhanced caption diversity and richness, we apply VisualFactChecker (VFC) \cite{ge_visual_2024}, an LLM-based technique for performing caption upsampling. We train all models for $800{,}000$ iterations at $256{\times}256$ resolution with a global batch size of $2{,}048$ on $32$ A100 GPUs. Each model has around $870$M parameters and takes approximately $7$ days to train.

\setlength{\tabcolsep}{3.5pt}
\begin{table*}[t]
\centering
\begin{tabular}{lccccccccccccc}
\toprule
\textbf{Model} & \textbf{Size} & \textbf{Avg} & \textbf{Attr.} & \textbf{Scene} & \textbf{Spat.} & \textbf{Action} & \textbf{Part} & \textbf{Count.} & \textbf{Comp.} & \textbf{Differ.} & \textbf{Neg.} & \textbf{Uni.}\\
\midrule
CLIP\textsubscript{ViT-H/14} & 354M & \C{0.622} & \C{0.612} & \C{0.731} & \C{0.608} & \C{0.655} & \C{0.594} & \C{0.529} & \C{0.522} & \C{0.425} & \C{0.480} & \C{0.632} \\
T5-XXL & 4.7B & \textbf{\C{0.741}} & \textbf{\C{0.737}} & \C{0.809} & \textbf{\C{0.741}} & \C{0.782} & \textbf{\C{0.723}} & \textbf{\C{0.677}} & \textbf{\C{0.717}} & \textbf{\C{0.675}} & \C{0.599} & \C{0.757} \\
\midrule
Qwen2-7B & 7B &\C{0.683} & \C{0.679} & \C{0.805} & \C{0.670} & \C{0.724} & \C{0.657} & \C{0.588} & \C{0.603} & \C{0.590} & \C{0.552} & \C{0.763} \\
Mistral-7B & 7B & \C{0.675} & \C{0.667} & \C{0.763} & \C{0.665} & \C{0.711} & \C{0.641} & \C{0.576} & \C{0.556} & \C{0.526} & \C{0.524} & \C{0.726} \\
Llama3-8B & 8B & \C{0.675} & \C{0.673} & \C{0.767} & \C{0.656} & \C{0.704} & \C{0.667} & \C{0.627} & \C{0.615} & \C{0.568} & \C{0.542} & \C{0.768} \\
Gemma2-9B & 9B & \C{0.710} & \C{0.709} & \C{0.794} & \C{0.711} & \C{0.760} & \C{0.705} & \C{0.642} & \C{0.659} & \C{0.617} & \C{0.544} & \C{0.709} \\
\midrule
gte-Qwen2 & 7B & \C{0.482} & \C{0.486} & \C{0.537} & \C{0.479} & \C{0.497} & \C{0.466} & \C{0.446} & \C{0.393} & \C{0.405} & \C{0.424} & \C{0.437} \\
sfr-Mistral & 7B & \C{0.710} & \C{0.706} & \C{0.804} & \C{0.707} & \C{0.740} & \C{0.691} & \C{0.661} & \C{0.670} & \C{0.615} & \C{0.608} & \C{0.766} \\
Mistral-7B\textsubscript{Instruct} & 7B & \C{0.690} & \C{0.683} & \C{0.787} & \C{0.686} & \C{0.718} & \C{0.654} & \C{0.628} & \C{0.630} & \C{0.589} & \C{0.577} & \C{0.762} \\
bge-Gemma2 & 9B & \C{0.737} & \C{0.730} & \textbf{\C{0.824}} & \C{0.729} & \textbf{\C{0.793}} & \C{0.722} & \C{0.662} & \C{0.654} & \C{0.641} & \textbf{\C{0.623}} & \textbf{\C{0.797}} \\
\bottomrule
\end{tabular}
\caption{VQAScore for models using embeddings extracted from the last layer. We use text encoders from \href{https://huggingface.co/laion/CLIP-ViT-H-14-laion2B-s32B-b79K}{CLIP-ViT-H/14} (354M) and \href{https://huggingface.co/google-t5/t5-11b}{T5-XXL} (4.7B), along with four popular open-source pre-trained LLMs: \href{https://huggingface.co/Qwen/Qwen2-7B}{Qwen2} (7B), \href{https://huggingface.co/mistralai/Mistral-7B-v0.1}{Mistral-7B} (7B), \href{https://huggingface.co/meta-llama/Meta-Llama-3-8B}{Llama3} (8B), and \href{https://huggingface.co/google/gemma-2-9b}{Gemma2} (9B). Additionally, we include three embedding models fine-tuned on these LLMs: gte-Qwen2 (\href{https://huggingface.co/Alibaba-NLP/gte-Qwen2-7B-instruct}{gte-Qwen2-7B-instruct}; 7B), sfr-Mistral (\href{https://huggingface.co/Salesforce/SFR-Embedding-2_R}{SFR-Embedding-2\_R}; 7B), and bge-Gemma2 (\href{https://huggingface.co/BAAI/bge-multilingual-gemma2}{bge-multilingual-gemma2}; 9B). We also include an instruction fine-tuned model, \href{https://huggingface.co/mistralai/Mistral-7B-Instruct-v0.2}{Mistral-7B-Instruct} (7B). The highest scores are shown in \textbf{bold}. Our results show that only using the last layer does not work well for LLMs and perform worse than T5.}
\label{tab:baseline}
\end{table*}
\setlength{\tabcolsep}{\oldtabcolsep}

\subsection{Models of interest} \label{sec:models}

We mainly explore four types of text encoders:
\begin{itemize}[topsep=2mm, itemsep=2mm]
    \item T5: \textbf{\href{https://huggingface.co/google-t5/t5-11b}{T5-XXL}} (encoder size $\approx4.7$B params) \cite{raffel_exploring_2023} is an encoder-decoder model that frames NLP tasks in a unified text-to-text format, utilizing a span-masked language model objective. Its text encoder is widely used in text-to-image models to effectively capture linguistic and semantic information for image generation.
    \item CLIP: \textbf{\href{https://huggingface.co/laion/CLIP-ViT-H-14-laion2B-s32B-b79K}{CLIP-ViT-H/14}} (encoder size $\approx354$M params) \cite{radford_learning_2021} aligns visual and textual representations within a shared embedding space through contrastive learning, making it a popular choice for text-to-image generation for strong text-image alignment.
    \item Pre-trained LLMs: \textbf{\href{https://huggingface.co/mistralai/Mistral-7B-v0.1}{Mistral-7B}} (and instruction fine-tuned \href{https://huggingface.co/mistralai/Mistral-7B-Instruct-v0.2}{Mistral-7B-Instruct}) \cite{jiang_mistral_2023}, \textbf{Gemma2} (\href{https://huggingface.co/google/gemma-2-2b}{2B} and \href{https://huggingface.co/google/gemma-2-9b}{9B}) \cite{team_gemma_2024}, \textbf{\href{https://huggingface.co/meta-llama/Meta-Llama-3-8B}{Llama3-8B}} \cite{dubey_llama_2024}, \textbf{Qwen2} (\href{https://huggingface.co/Qwen/Qwen2-1.5B}{1.5B} and \href{https://huggingface.co/Qwen/Qwen2-7B}{7B}) \cite{yang_qwen2_2024}. We include several open-source, high-performing LLMs in our study, each trained for autoregressive language modeling with a next token prediction objective. Although these models are all decoder-only transformers, they vary in terms of tokenization and other architectural details.
    \item LLM fine-tuned embedding models: \textbf{bge-Gemma2} (\href{https://huggingface.co/BAAI/bge-multilingual-gemma2}{bge-multilingual-gemma2}; 9B) \cite{chen_bge_2024}, \textbf{gte-Qwen2} (\href{https://huggingface.co/Alibaba-NLP/gte-Qwen2-7B-instruct}{gte-Qwen2-7B-instruct}; 7B) \cite{li_towards_2023}, \textbf{sfr-Mistral} (\href{https://huggingface.co/Salesforce/SFR-Embedding-2_R}{SFR-Embedding-2\_R}; 7B) \cite{sfr-embed}. We also evaluate fine-tuned versions of the LLMs mentioned above, specifically selected from top performing models on the Massive Text Embedding Benchmark (MTEB) Leaderboard\footnote{All ranked top 10 on the MTEB Leaderboard as of Nov. 2024.} \cite{muennighoff_mteb_2023}. We aim to examine whether improvements in semantic comprehension translate into improved image generation and text--image alignment.
\end{itemize}

\subsection{Benchmarking and metrics}

To evaluate and compare the performance of different models, we adopt GenAI-Bench \cite{lin_evaluating_2024} as our primary benchmarking suite. GenAI-Bench includes $1{,}600$ diverse and challenging prompts, and each prompt is annotated with specific aspects in the compositional text-to-visual generation: \textit{Attribute}, \textit{Scene}, \textit{Spatial Relation}, \textit{Action Relation}, \textit{Part Relation}, \textit{Counting}, \textit{Differentiation}, \textit{Comparison}, \textit{Negation}, and \textit{Universality} \cite{lin_evaluating_2024, li_genai-bench_2024}. These skill annotations enable us to conduct in-depth ablation studies, allowing for detailed analysis of how various LLMs and embedding extraction methods affect specific aspects of text-to-image generation. To better align the prompts with our upsampled training distribution, we utilize Gemma2-9B \cite{team_gemma_2024} for prompt upsampling as detailed in the supplementary material.

\begin{figure}[t]
\includegraphics[scale=0.44]{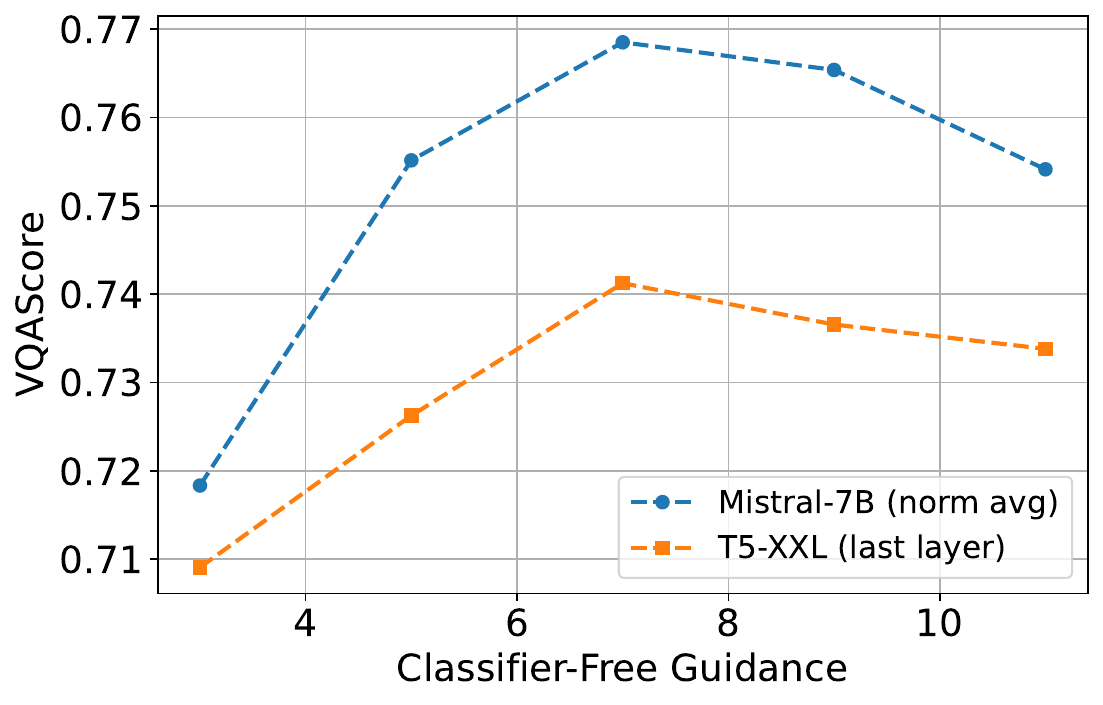}
\centering
\caption{VQAScore as a function of classifier-free guidance weight. We show the VQAScore of Mistral-7B using layer-normalized average embeddings (\textit{blue}) and T5-XXL using last-layer embeddings (\textit{orange}) at varying guidance strengths.}
\label{fig:cfg_plot}
\end{figure}

\setlength{\tabcolsep}{5.3pt}
\begin{table*}[t]
\centering
\small
\begin{tabular}{l@{\hspace{3mm}}lcccccccccccc}
\toprule
\textbf{Model} & \textbf{Layer} & \textbf{Avg} & \textbf{Attr.} & \textbf{Scene} & \textbf{Spat.} & \textbf{Action} & \textbf{Part} & \textbf{Count.} & \textbf{Comp.} & \textbf{Differ.} & \textbf{Neg.} & \textbf{Uni.}\\
\midrule
T5-XXL & 25 (last) & \textbf{\C{0.741}} & \textbf{\C{0.737}} & \C{0.809} & \textbf{\C{0.741}} & \textbf{\C{0.782}} & \textbf{\C{0.723}} & \textbf{\C{0.677}} & \textbf{\C{0.717}} & \textbf{\C{0.675}} & \textbf{\C{0.599}} & \textbf{\C{0.757}} \\
Mistral-7B & 33 (last) & \C{0.675} & \C{0.667} & \C{0.763} & \C{0.665} & \C{0.711} & \C{0.641} & \C{0.576} & \C{0.556} & \C{0.526} & \C{0.524} & \C{0.726} \\
Mistral-7B & 32 & \C{0.710} & \C{0.718} & \C{0.794} & \C{0.707} & \C{0.747} & \C{0.686} & \C{0.638} & \C{0.625} & \C{0.650} & \C{0.582} & \C{0.703} \\
Mistral-7B & 15 & \C{0.725} & \C{0.720} & \textbf{\C{0.812}} & \C{0.729} & \C{0.762} & \C{0.692} & \C{0.673} & \C{0.651} & \C{0.647} & \C{0.585} & \C{0.732} \\
Mistral-7B & 0\hphantom{0} (first) & \C{0.375} & \C{0.372} & \C{0.499} & \C{0.339} & \C{0.379} & \C{0.290} & \C{0.297} & \C{0.328} & \C{0.200} & \C{0.278} & \C{0.440} \\
\bottomrule
\end{tabular}
\caption{VQAScore for models using embeddings extracted from individual layers of \href{https://huggingface.co/mistralai/Mistral-7B-v0.1}{Mistral-7B} (7B). We include the baseline \href{https://huggingface.co/google-t5/t5-11b}{T5-XXL} (4.7B) model using embeddings extracted from the last layer as a reference. The highest scores are shown in \textbf{bold}. Our results show that using middle layers can outperform earlier or later layers.}
\label{tab:individualLayers}
\end{table*}
\setlength{\tabcolsep}{\oldtabcolsep}
\setlength{\tabcolsep}{4.2pt}
\begin{table*}[t]
\centering
\small
\begin{tabular}{lccccccccccccc}
\toprule
\textbf{Model} & \textbf{Embeddings} & \textbf{Avg} & \textbf{Attr.} & \textbf{Scene} & \textbf{Spat.} & \textbf{Action} & \textbf{Part} & \textbf{Count.} & \textbf{Comp.} & \textbf{Differ.} & \textbf{Neg.} & \textbf{Uni.}\\
\midrule
T5-XXL & last layer & \C{0.741} & \C{0.737} & \C{0.809} & \C{0.741} & \C{0.782} & \C{0.723} & \C{0.677} & \C{0.717} & \C{0.675} & \C{0.599} & \C{0.757} \\
T5-XXL & norm avg & \C{0.747} & \C{0.748} & \C{0.813} & \C{0.745} & \C{0.780} & \C{0.720} & \C{0.687} & \C{0.736} & \C{0.675} & \C{0.617} & \C{0.760} \\
\midrule
bge-Gemma2 & last layer & \C{0.737} & \C{0.730} & \C{0.824} & \C{0.729} & \C{0.793} & \C{0.722} & \C{0.662} & \C{0.654} & \C{0.641} & \C{0.623} & \C{0.797} \\
bge-Gemma2 & avg & \C{0.774} & \C{0.776} & \textbf{\C{0.851}} & \C{0.765} & \C{0.813} & \C{0.752} & \C{0.716} & \C{0.773} & \C{0.702} & \C{0.665} & \textbf{\C{0.822}} \\
bge-Gemma2 & norm avg & \textbf{\C{0.789}} & \textbf{\C{0.787}} & \C{0.846} & \textbf{\C{0.782}} & \textbf{\C{0.821}} & \textbf{\C{0.786}} & \textbf{\C{0.745}} & \textbf{\C{0.776}} & \textbf{\C{0.744}} & \textbf{\C{0.712}} & \C{0.810} \\
\midrule
Mistral-7B & last layer & \C{0.675} & \C{0.667} & \C{0.763} & \C{0.665} & \C{0.711} & \C{0.641} & \C{0.576} & \C{0.556} & \C{0.526} & \C{0.524} & \C{0.726} \\
Mistral-7B & avg & \C{0.731} & \C{0.733} & \C{0.806} & \C{0.731} & \C{0.779} & \C{0.712} & \C{0.683} & \C{0.666} & \C{0.639} & \C{0.582} & \C{0.758} \\
Mistral-7B & norm avg & \C{0.769} & \C{0.774} & \C{0.837} & \C{0.780} & \C{0.802} & \C{0.733} & \C{0.699} & \C{0.716} & \C{0.706} & \C{0.630} & \C{0.789} \\
\bottomrule
\end{tabular}
\caption{VQAScore for models using different embedding strategies: standard last-layer embeddings (\textit{last layer}), average embeddings across all layers (\textit{avg}), and average embeddings across all normalized layers (\textit{norm avg}). We apply these approaches to the encoder from the baseline \href{https://huggingface.co/google-t5/t5-11b}{T5-XXL} (4.7B), the pre-trained LLM \href{https://huggingface.co/mistralai/Mistral-7B-v0.1}{Mistral-7B} (7B), and our best performing fine-tuned embedding model bge-Gemma2 (\href{https://huggingface.co/BAAI/bge-multilingual-gemma2}{bge-multilingual-gemma2}; 9B). The highest scores are shown in \textbf{bold}. Our results show that using layer-normalized averaging greatly enhances performance and outperforms T5.}
\label{tab:averaging}
\end{table*}
\setlength{\tabcolsep}{\oldtabcolsep}

Previous research has shown that traditional text-image evaluation metrics like CLIPScore \cite{hessel_clipscore_2022} and FID \cite{heusel_gans_2018} are not consistent with human evaluations, particularly in assessing complex compositional tasks \cite{lin_evaluating_2024, li_genai-bench_2024, imagen-team-google_imagen_2024}. Instead, VQA-based automatic evaluation methods have demonstrated higher reliability and correlation with human judgements \cite{wiles_revisiting_2024, hu_tifa_2023, huang_t2i-compbench_2023, yarom_what_2023, cho_visual_2023, li_genai-bench_2024, lin_evaluating_2024}. Our observations are consistent with these findings, so we use VQAScore \cite{lin_evaluating_2024} as our primary evaluation metric in this work. Notably, we also observe that the custom CLIP-FlanT5 \cite{lin_evaluating_2024} model that was originally proposed for VQAScore is not sufficient to differentiate between similar models \cite{imagen-team-google_imagen_2024}. To address this, we utilize GPT-4o \cite{openai_gpt-4o_2024} for our implementation of VQAScore, achieving improved differentiation between models and a closer match to human-perceived quality. When computing VQAScore, we generate the $1{,}600$ images using the same seeds for each model. The random variation of VQAScore with GPT-4o is up to $\pm0.003$, depending on the category. Please find more details in the supplementary material.

\subsection{Classifier-free guidance}
In our work, we use a classifier-free guidance \cite{ho2022classifierfreediffusionguidance} of $7.0$ for all models. Figure \ref{fig:cfg_plot} shows a line graph with VQAScore as a function of guidance weight. These results illustrate how different guidance strength affects VQAScore considerably, potentially even more than the choice of the text encoder. As such, we standardize guidance at $7.0$ in our evaluation pipeline for apples-to-apples comparison. Similar to the original Stable Diffusion model \cite{Rombach_2022_CVPR}, we do not employ any kind of thresholding \cite{saharia_photorealistic_2022} or rescaling \cite{lin_common_2024}.

\section{Results}
\label{sec:results}

\subsection{Final layer LLM embeddings lack in visio-linguistic reasoning} \label{sec:logic}
We begin by training text-to-image models using text encoders described in Sec. \ref{sec:models}, extracting embeddings from the final layer, as commonly done in the literature~\cite{saharia_photorealistic_2022, gao2024lumina, zhuo2024lumina, xie2024sana}.
As shown in Table \ref{tab:baseline}, replacing T5 with other LLMs consistently results in a decrease in performance. Most of the performance gap can be attributed to the advanced visio-linguistic reasoning skills in GenAI-Bench: \textit{Counting}, \textit{Differentiation}, \textit{Comparison}, \textit{Negation}, and \textit{Universality}. Specifically, \textit{Comparison} shows the most significant drop in performance relative to T5. 
In fact, when prompts tagged with \textit{Comparison} are excluded, the results for the LLMs are similar to or even better than T5. 

On the other hand, CLIP demonstrates one of the lowest scores overall, likely due to a combination of its much smaller model size, shorter token sequence length of $77$, and reduced ability to capture linguistic and semantic details compared to the other language models \cite{radford_learning_2021, thrush_winoground_2022, castro_scalable_2023}. For these reasons, we use T5 as the primary baseline in subsequent analysis.

\setlength{\tabcolsep}{4.5pt}
\begin{table*}[t]
\centering
\begin{tabular}{lcccccccccccc}
\toprule
\textbf{Model} & \textbf{Avg} & \textbf{Attr.} & \textbf{Scene} & \textbf{Spat.} & \textbf{Action} & \textbf{Part} & \textbf{Count.} & \textbf{Comp.} & \textbf{Differ.} & \textbf{Neg.} & \textbf{Uni.}\\
\midrule
bge-Gemma2 & \C{0.737} & \C{0.730} & \C{0.824} & \C{0.729} & \C{0.793} & \C{0.722} & \C{0.662} & \C{0.654} & \C{0.641} & \C{0.623} & \C{0.797} \\
bge-Gemma2\textsubscript{pooled} & \C{0.737} & \C{0.733} & \C{0.838} & \C{0.725} & \C{0.767} & \C{0.712} & \C{0.692} & \C{0.691} & \C{0.663} & \C{0.641} & \C{0.823} \\
\midrule
sfr-Mistral & \C{0.710} & \C{0.706} & \C{0.804} & \C{0.707} & \C{0.740} & \C{0.691} & \C{0.661} & \C{0.670} & \C{0.615} & \C{0.608} & \C{0.766} \\
sfr-Mistral\textsubscript{pooled} & \C{0.698} & \C{0.685} & \C{0.810} & \C{0.693} & \C{0.748} & \C{0.694} & \C{0.610} & \C{0.641} & \C{0.596} & \C{0.589} & \C{0.730} \\
\bottomrule
\end{tabular}
\caption{VQAScore for models using embeddings extracted from the last layer compared to models with additional conditioning on global pooled embeddings \cite{podell_sdxl_2023}. We evaluate the fine-tuned embedding models bge (\href{https://huggingface.co/BAAI/bge-multilingual-gemma2}{bge-multilingual-gemma2}; 9B) and sfr (\href{https://huggingface.co/Salesforce/SFR-Embedding-2_R}{SFR-Embedding-2\_R}; 7B). We observe limited differences with the latter approach for embedding models.}
\label{tab:pooling}
\end{table*}
\setlength{\tabcolsep}{\oldtabcolsep}
\setlength{\tabcolsep}{4.8pt}
\begin{table*}[ht]
\centering
\begin{tabular}{lccccccccccccc}
\toprule
\textbf{Model} & \textbf{Size} & \textbf{Avg} & \textbf{Attr.} & \textbf{Scene} & \textbf{Spat.} & \textbf{Action} & \textbf{Part} & \textbf{Count.} & \textbf{Comp.} & \textbf{Differ.} & \textbf{Neg.} & \textbf{Uni.}\\
\midrule
Qwen2 & 1.5B & \C{0.655} & \C{0.654} & \C{0.758} & \C{0.642} & \C{0.681} & \C{0.624} & \C{0.575} & \C{0.553} & \C{0.528} & \C{0.553} & \C{0.676} \\
Qwen2 & 7B & \C{0.683} & \C{0.679} & \C{0.805} & \C{0.670} & \C{0.724} & \C{0.657} & \C{0.588} & \C{0.603} & \C{0.590} & \C{0.552} & \C{0.763} \\
\midrule
Gemma2 & 2B & \C{0.640} & \C{0.650} & \C{0.745} & \C{0.627} & \C{0.687} & \C{0.606} & \C{0.580} & \C{0.597} & \C{0.527} & \C{0.442} & \C{0.707} \\
Gemma2 & 9B & \C{0.710} & \C{0.709} & \C{0.794} & \C{0.711} & \C{0.760} & \C{0.705} & \C{0.642} & \C{0.659} & \C{0.617} & \C{0.544} & \C{0.709} \\
\bottomrule
\end{tabular}
\caption{VQAScore for models with different LLM sizes, using embeddings extracted from the last layer. We evaluate the pre-trained LLMs: \href{https://huggingface.co/Qwen/Qwen2-1.5B}{Qwen2} (1.5B), \href{https://huggingface.co/Qwen/Qwen2-7B}{Qwen2} (7B), \href{https://huggingface.co/google/gemma-2-2b}{Gemma2} (2B), \href{https://huggingface.co/google/gemma-2-9b}{Gemma2} (9B). Our results show that LLM scaling improves performance but does not uniformly impact all aspects of image composition.}
\label{tab:modelSizes}
\end{table*}
\setlength{\tabcolsep}{\oldtabcolsep}

\subsection{Embedding performance differs by layer}
In addition to using the final layer, we experiment with embeddings extracted from other individual layers within the Mistral-7B model to assess how semantic and linguistic representations across layers vary and impact image generation. The Mistral model, with its 33 layers, has been the focus of several works exploring embedding representations\footnote{Llama3, another popular LLM, has \href{https://huggingface.co/astronomer/Llama-3-8B-Special-Tokens-Adjusted}{untrained token embedding issues}} \cite{jiang_mistral_2023, behnamghader_llm2vec_2024, springer_repetition_2024, wang_improving_2024}. Table \ref{tab:individualLayers} shows that using only the $15$th layer of Mistral outperforms the use of the $33$rd (last), $0$th (first), and $32$nd (penultimate) layers. This suggests that different layers of LLMs capture distinct levels of semantic and linguistic understanding \cite{dar2023analyzingtransformersembeddingspace}. Early layers may primarily encode basic linguistic structures, while later layers tend to be specialized in next-token prediction \cite{liu_fantastic_2024}. By contrast, middle layers appear to offer a more balanced abstraction of semantic representations, which can better support image generation tasks. However, the baseline last layer T5 model still outperforms any single layer Mistral model. This leads to a key insight: \textbf{using embeddings from any single layer of the LLMs for text-to-image models is not sufficient}.

\subsection{Averaging layers yields stronger embeddings}
\label{sec:averaging}
Since different layers in LLMs encode varying levels of semantic representations, we propose to aggregate the embeddings using two approaches. The first method involves directly averaging the embeddings across all layers. In the second approach, we apply mean normalization to each layer’s embeddings before averaging. This ensures that each layer contributes consistently in scale, preventing any single layer from disproportionately influencing the aggregated representation \cite{zhang_investigating_2024}. We apply these approaches to Mistral and bge-Gemma2, our best performing embedding model. We show our results in Table \ref{tab:averaging} and Figure \ref{fig:radar}. More results with other models, such as Qwen2 and Llama3, can be found in supplementary material Table \ref{tab:norm_avg_models}. For both models, averaged embeddings yield substantial improvements over using only the last layer, even surpassing the baseline T5 model by a large margin. Notably, these approaches overcome the previously observed limitations in advanced visio-linguistic reasoning, demonstrating that embeddings derived from all layers produce richer and more comprehensive representations. 

Our experiments also reveal that applying layer normalization beforehand performs even better than only averaging, as it balances contributions from each layer and prevents any single layer from dominating the representation. This leads to another key finding: \textbf{the most effective way to utilize LLMs for text-to-image models is to properly use embeddings from all layers.}



Interestingly, applying layer-normalized averaging to T5 embeddings shows almost no performance difference. We believe this is due to the T5 encoder's approach of framing each task in a text-to-text format, and its later layers inherently build upon the semantic and linguistic foundations established in earlier layers. In contrast, the later layers of decoder-only LLMs tend to specialize in next-token prediction, which explains the significant performance gains observed when utilizing embeddings from intermediate layers.

\setlength{\tabcolsep}{1pt}
\renewcommand{\arraystretch}{0.5}
\begin{figure*}[t]
    \centering
    \begin{tabular}{cccc}
        {\centering Mistral (\emph{last layer})} & {\centering Mistral (\emph{norm avg})} & {\centering Mistral (\emph{last layer})} & {\centering Mistral (\emph{norm avg})} \\[2mm] 
        \includegraphics[width=0.245\textwidth]{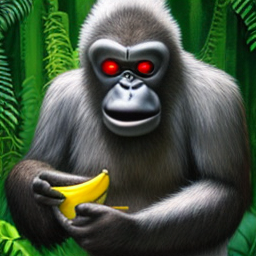} &
        \includegraphics[width=0.245\textwidth]{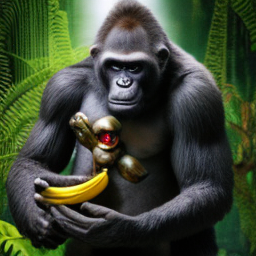} &
        \includegraphics[width=0.245\textwidth]{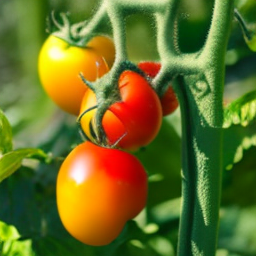} &
        \includegraphics[width=0.245\textwidth]{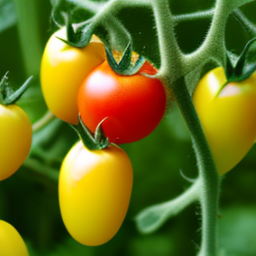} \\
        \includegraphics[trim=8 8 8 8, clip, width=0.245\textwidth]{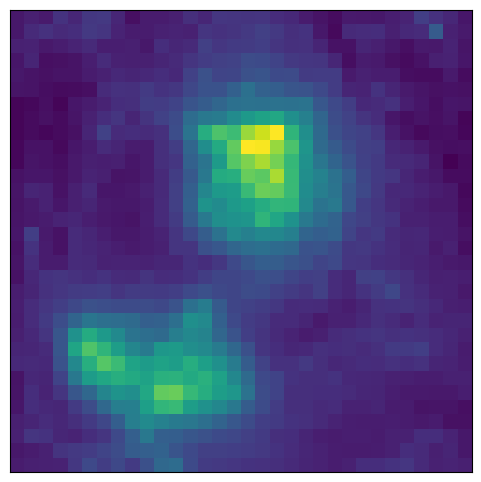} &
        \includegraphics[trim=8 8 8 8, clip, width=0.245\textwidth]{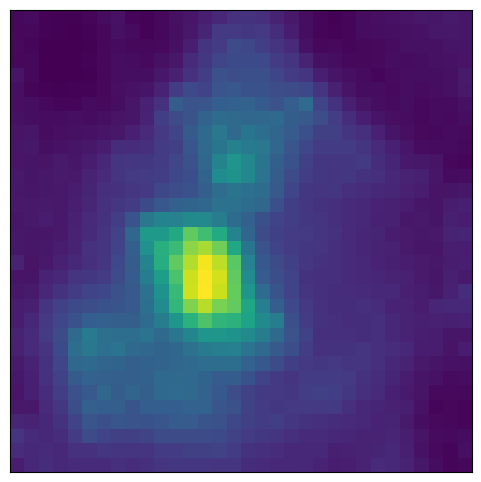} &
        \includegraphics[trim=8 8 8 8, clip, width=0.245\textwidth]{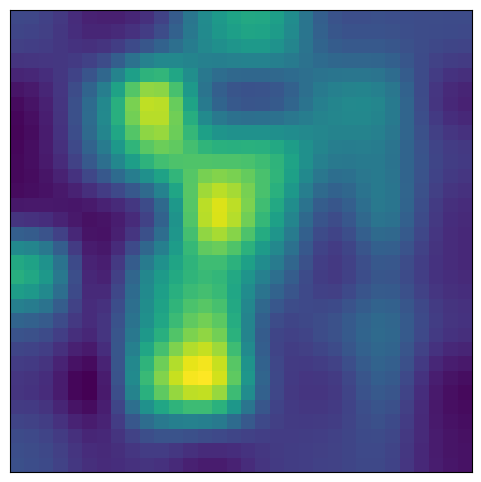} &
        \includegraphics[trim=8 8 8 8, clip, width=0.245\textwidth]{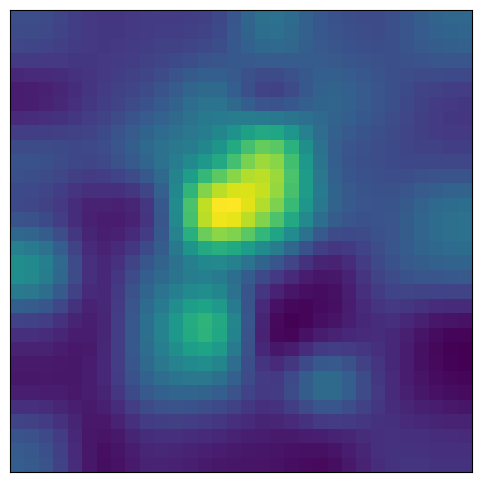} \\ [2mm]

        \multicolumn{2}{c}{\emph{A larger gorilla hands
a smaller }} & \multicolumn{2}{c}{\emph{A tomato vine with several tomatoes on it,}} \\[1mm]
        \multicolumn{2}{c}{\emph{mechanical \textbf{monkey} a banana}} & \multicolumn{2}{c}{\emph{all yellow except the \textbf{largest} which is red}} \\[2mm]
\end{tabular}
\caption{Heatmap visualization. Images generated with Mistral-7B using standard last-layer embeddings (\textit{last layer}) and layer-normalized average embeddings (\textit{norm avg}). Corresponding cross-attention heatmaps for the tokens \textbf{monkey} (left set) and \textbf{largest} (right set) are shown in the panel below. These visualizations show how the \textit{norm avg} model performs better than \textit{last layer} model on prompts that require advanced visio-linguistic reasoning skills such as \textit{differentiation} (left set) and \textit{comparison} (right set).}
\label{fig:heatmap}
\end{figure*}
\renewcommand{\arraystretch}{1}
\setlength{\tabcolsep}{\oldtabcolsep}

\subsection{Finetuned embedding models show potential}
Our results in Table \ref{tab:baseline} indicate that fine-tuned embedding models can offer improvement over their pre-trained LLM counterparts, as seen with bge-Gemma2 and sfr-Mistral, but may also lead to significant performance degradation, as with gte-Qwen2. We suspect this inconsistency arises because most fine-tuned embedding models are optimized to produce pooled embeddings primarily through methods like last token pooling, mean pooling, or weighted mean pooling. These pooling strategies may impact the embeddings of individual tokens, resulting in less meaningful representations across the entire sequence \cite{springer_repetition_2024, lee_nv-embed_2024}. 

We also experiment with a SDXL \cite{podell_sdxl_2023} inspired approach by conditioning the text-to-image models on two types of embeddings: standard per-token embeddings and a single pooled text embedding as a condensed global representation. Individual tokens go to cross-attention, while the pooled token is combined with the timestep embedding to modulate layer activations. Since these embedding models are primarily designed to produce a single pooled embedding, this approach enables us to incorporate their intended use case fully for better comparison to our baseline models. We selected bge-Gemma2 and sfr-Mistral for their improvement over their base models. However, as shown in Table \ref{tab:pooling}, we observe limited differences in performance with this method, suggesting that our approach may not fully leverage the potential of these embeddings.

Despite these mixed results, \textbf{there is considerable promise in using embedding models for text-to-image generation}, as demonstrated by the bge-Gemma2 model with layer-normalized averaging, which is the best performing model in our experiments. Future research to determine the most effective strategies could explore hybrid pooling approaches that combine per-token embeddings from pre-trained LLMs with additional conditioning from pooled embeddings in fine-tuned embedding models. Adding an extra latent attention layer, proposed by NV-Embed \cite{lee_nv-embed_2024}, could also improve the quality of pooled embeddings.

\subsection{Scaling improves performance, but not uniformly across compositional skills}
Prior research has shown that scaling the size of the text encoder yields greater improvements in image quality and text-image alignment compared to increasing the size of the diffusion model \cite{saharia_photorealistic_2022}. In our experiments, we evaluate different LLM model sizes to understand the impact of scaling on performance. Our results on differently sized Qwen2 and Gemma2 models in Table \ref{tab:modelSizes} confirm these findings. Although our ablation experiments are limited to a subset of models, the skill annotations from GenAI-Bench reveal an interesting insight: \textbf{while performance increases with model size, scaling does not uniformly impact all aspects of compositional text-to-visual generation}. For instance, we observe minimal improvement in \textit{Negation} and \textit{Counting} between the two Qwen2 models, while the difference in \textit{Universality} is significant. Conversely, the two Gemma2 models show little difference in \textit{Universality} but exhibit substantial differences across most other skills.

Given the minimal improvements in specific skills, these results raise questions about the efficiency of simply scaling model size for text-to-image generation. Future work could explore hybrid approaches, such as combining different LLMs to leverage complementary strengths. This is similar to how eDiff-I \cite{balaji_ediff-i_2023} combined the T5 and CLIP text encoders to enhance embedding representations. Smaller, specialized models fine-tuned on skill-specific tasks might also provide a more efficient way to improve performance. 

Additionally, it would be valuable to investigate the effects of scaling text encoders when fully utilizing all layers, such as our approaches using layer-normalized averaging. This could reveal whether more refined approaches to layer utilization might achieve comparable or even superior results to model scaling.
\section{Visual analysis}
\label{sec:analysis}


As shown in Sections~\ref{sec:logic} and \ref{sec:averaging}, LLMs demonstrate significant improvements in advanced visio-linguistic reasoning when using layer-normalized average embeddings. Figure \ref{fig:visual_comparison} provides a visual comparison with T5 and CLIP models for handling \textit{negation}, where the baselines fail to interpret such logic and incorrectly depict the dog wearing clothes. Additional comparisons are available in the supplementary material. We observe these enhanced logical abilities in our models with Mistral through visual comparisons and cross-attention heatmaps on relevant tokens:

\textbf{Differentiation.}
The left set of images in Figure \ref{fig:heatmap} illustrate a \textit{differentiation} prompt involving a gorilla and a mechanical monkey. The model using last-layer embeddings fails to distinguish between the two, rendering the gorilla with robotic features but omitting the monkey. Cross-attention heatmaps for the token \textit{monkey} show that attention is focused on the gorilla, indicating a lack of differentiation.

In contrast, the model using averaged layer-normalized embeddings successfully separates the two entities. The cross-attention heatmap for monkey shows distinct attention away from the gorilla, allowing the mechanical monkey to appear separately. This suggests that layer-normalized averaging helps the model better differentiate semantically similar objects, as seen in the increased embedding norm distinction between gorilla and monkey. Additionally, we observe that the difference in token embedding norms for \textit{gorilla} and \textit{monkey} is greater in layer-normalized averaging compared to the last layer. This increased distinction in embedding norms likely contributes to the model’s improved differentiation ability.

\textbf{Comparison.} 
The right set of images in Figure \ref{fig:heatmap} show results for \textit{comparison} prompts, where only the largest tomato should be red. With last-layer embeddings, multiple tomatoes appear red, indicating the model’s struggle to interpret logical constraints. Cross-attention heatmaps for \textit{largest} show diffused attention, which contributes to misalignment with the prompt.

In contrast, layer-normalized average embeddings lead to better alignment with the prompt, focusing attention on the largest tomato and accurately rendering it as red while keeping the others yellow. However, neither model successfully represents the largest tomato as larger than the others, suggesting that further research is needed to explore how size-related cues are embedded and translated to image generation for improved spatial and comparative alignment.
\section{Conclusion}
\label{sec:conclusion}
Our research contributes to the evolving field of text-to-image generation by exploring the use of decoder-only LLMs as the text encoder. Through a controlled training pipeline, we isolate and evaluate the impact of different text encoders, conducting extensive experiments on various LLMs and embedding extraction methods.

Our findings provide valuable insights into using decoder-only LLMs for text-to-image generation. We demonstrate that relying on a single layer from an LLM is insufficient and propose a simple yet effective approach that leverages all layers through averaged layer-normalized embeddings. With this approach, our models exhibit significant improvements in advanced visio-linguistic reasoning skills, outperforming the baseline T5 model across all aspects of compositional text-to-image generation. Furthermore, our experiments with fine-tuned embedding models reveal promising potential for enhanced contextual semantic comprehension. Additionally, we explore the scalability of LLMs, observing that while performance generally increases with model size, specific compositional skills do not uniformly benefit from scaling.

This work opens several avenues for future research, including more refined approaches to layer aggregation, optimized usage of embedding models, and the development of hybrid models that combine multiple LLMs. By providing a detailed examination of decoder-only LLMs in text-to-image generation, we hope our insights will guide future advancements and extend to other multimodal applications.
\clearpage
\small
\bibliographystyle{ieeenat_fullname}
\bibliography{ref}

\clearpage
\setcounter{page}{1}
\maketitlesupplementary

\setcounter{section}{0}
\renewcommand*{\thesection}{\Alph{section}}

\setcounter{table}{0}
\renewcommand*{\thetable}{\Alph{table}}

\setcounter{figure}{0}
\renewcommand*{\thefigure}{\Alph{figure}}

\begin{table*}[t]
\centering
\begin{tabular}{ccc}
\toprule
\textbf{Text Encoders} & \textbf{Embedding Dimension} & \textbf{Extra Parameters} \\
CLIP, T5 & $1024$ & $1{,}048{,}576$ \\
Qwen2-1.5B  & $1536$ & $1{,}572{,}864$ \\
Gemma2-2B & $2304$ & $2{,}359{,}296$ \\
Qwen2-7B, Gemma2-9B, gte-Gwen2, bge-Gemma2 & $3584$ & $3{,}670{,}016$ \\
Mistral-7B, Llama3-8B, sfr-Mistral, Mistral-Instruct & $4096$ & $4{,}194{,}304$ \\
\bottomrule
\end{tabular}
\caption{Additional trainable parameters from adding a linear projection layer from text encoder’s embedding dimensions to $1024$ output features before cross-attention.}
\label{tab:emb_dims}
\end{table*}

\setlength{\tabcolsep}{4.2pt}
\begin{table*}[t]
\centering
\small
\begin{tabular}{lccccccccccccc}
\toprule
\textbf{Model} & \textbf{Embeddings} & \textbf{Avg} & \textbf{Attr.} & \textbf{Scene} & \textbf{Spat.} & \textbf{Action} & \textbf{Part} & \textbf{Count.} & \textbf{Comp.} & \textbf{Differ.} & \textbf{Neg.} & \textbf{Uni.}\\
\midrule
T5-XXL & last layer & \C{0.741} & \C{0.737} & \C{0.809} & \C{0.741} & \C{0.782} & \C{0.723} & \C{0.677} & \C{0.717} & \C{0.675} & \C{0.599} & \C{0.757} \\
T5-XXL & norm avg & \C{0.747} & \C{0.748} & \C{0.813} & \C{0.745} & \C{0.780} & \C{0.720} & \C{0.687} & \C{0.736} & \C{0.675} & \C{0.617} & \C{0.760} \\
\midrule
Qwen2-7B & last layer &\C{0.683} & \C{0.679} & \C{0.805} & \C{0.670} & \C{0.724} & \C{0.657} & \C{0.588} & \C{0.603} & \C{0.590} & \C{0.552} & \C{0.763} \\
Qwen2-7B & norm avg & \C{0.740} & \C{0.741} & \C{0.823} & \C{0.740} & \C{0.772} & \C{0.731} & \C{0.680} & \C{0.704} & \C{0.683} & \C{0.589} & \C{0.739} \\
\midrule
Mistral-7B & last layer & \C{0.675} & \C{0.667} & \C{0.763} & \C{0.665} & \C{0.711} & \C{0.641} & \C{0.576} & \C{0.556} & \C{0.526} & \C{0.524} & \C{0.726} \\
Mistral-7B & norm avg & \C{0.769} & \C{0.774} & \C{0.837} & \C{0.780} & \C{0.802} & \C{0.733} & \C{0.699} & \C{0.716} & \C{0.706} & \C{0.630} & \C{0.789} \\
\midrule
Llama3-8B & last layer & \C{0.675} & \C{0.673} & \C{0.767} & \C{0.656} & \C{0.704} & \C{0.667} & \C{0.627} & \C{0.615} & \C{0.568} & \C{0.542} & \C{0.768} \\
Llama3-8B & norm avg & \C{0.744} & \C{0.744} & \C{0.831} & \C{0.744} & \C{0.783} & \C{0.705} & \C{0.704} & \C{0.675} & \C{0.659} & \C{0.628} & \C{0.782} \\
\midrule
Gemma2-9B & last layer & \C{0.710} & \C{0.709} & \C{0.794} & \C{0.711} & \C{0.760} & \C{0.705} & \C{0.642} & \C{0.659} & \C{0.617} & \C{0.544} & \C{0.709} \\
Gemma2-9B & norm avg & \C{0.753} & \C{0.757} & \C{0.814} & \C{0.743} & \C{0.790} & \C{0.735} & \C{0.691} & \C{0.703} & \C{0.679} & \C{0.651} & \C{0.770} \\
\midrule
gte-Qwen2 & last layer & \C{0.482} & \C{0.486} & \C{0.537} & \C{0.479} & \C{0.497} & \C{0.466} & \C{0.446} & \C{0.393} & \C{0.405} & \C{0.424} & \C{0.437} \\
gte-Qwen2 & norm avg & \C{0.654} & \C{0.647} & \C{0.746} & \C{0.626} & \C{0.696} & \C{0.632} & \C{0.539} & \C{0.619} & \C{0.536} & \C{0.538} & \C{0.683} \\
\midrule
sfr-Mistral & last layer & \C{0.710} & \C{0.706} & \C{0.804} & \C{0.707} & \C{0.740} & \C{0.691} & \C{0.661} & \C{0.670} & \C{0.615} & \C{0.608} & \C{0.766} \\
sfr-Mistral & norm avg & \C{0.750} & \C{0.745} & \C{0.839} & \C{0.762} & \C{0.782} & \C{0.713} & \C{0.677} & \C{0.715} & \C{0.706} & \C{0.610} & \C{0.785} \\
\midrule
bge-Gemma2 & last layer & \C{0.737} & \C{0.730} & \C{0.824} & \C{0.729} & \C{0.793} & \C{0.722} & \C{0.662} & \C{0.654} & \C{0.641} & \C{0.623} & \C{0.797} \\
bge-Gemma2 & norm avg & \textbf{\C{0.789}} & \textbf{\C{0.787}} & \textbf{\C{0.846}} & \textbf{\C{0.782}} & \textbf{\C{0.821}} & \textbf{\C{0.786}} & \textbf{\C{0.745}} & \textbf{\C{0.776}} & \textbf{\C{0.744}} & \textbf{\C{0.712}} & \textbf{\C{0.810}} \\
\bottomrule
\end{tabular}
\caption{VQAScore for models using different embedding strategies: standard last-layer embeddings (last layer) and average embeddings across all normalized layers (norm avg). Highest scores are shown in \textbf{bold}. Our results show that using layer-normalized averaging significantly enhances performance and most models outperform T5.}
\label{tab:norm_avg_models}
\end{table*}
\setlength{\tabcolsep}{\oldtabcolsep}
\setlength{\tabcolsep}{3pt}
\begin{table*}
\centering
\begin{tabular}{lccccccccccccc}
\toprule
\textbf{Model} & \textbf{Size} & \textbf{Avg} & \textbf{Attr.} & \textbf{Scene} & \textbf{Spat.} & \textbf{Action} & \textbf{Part} & \textbf{Count.} & \textbf{Comp.} & \textbf{Differ.} & \textbf{Neg.} & \textbf{Uni.}\\
\midrule
CLIP\textsubscript{ViT-H/14} & 354M & \C{0.761} & \C{0.762} & \C{0.790} & \C{0.765} & \C{0.762} & \C{0.749} & \C{0.755} & \C{0.758} & \C{0.729} & \C{0.710} & \C{0.771} \\
T5-XXL & 4.7B & \C{0.795} & \C{0.795} & \C{0.816} & \C{0.803} & \C{0.803} & \C{0.780} & \C{0.799} & \C{0.800} & \C{0.793} & \C{0.739} & \C{0.804} \\
\midrule
Qwen2-7B & 7B & \C{0.772} & \C{0.772} & \C{0.798} & \C{0.777} & \C{0.782} & \C{0.761} & \C{0.760} & \C{0.759} & \C{0.748} & \C{0.712} & \C{0.785} \\
Mistral-7B & 7B & \C{0.767} & \C{0.765} & \C{0.787} & \C{0.771} & \C{0.771} & \C{0.751} & \C{0.756} & \C{0.753} & \C{0.733} & \C{0.710} & \C{0.780} \\
Llama3-8B & 8B & \C{0.770} & \C{0.769} & \C{0.795} & \C{0.773} & \C{0.775} & \C{0.767} & \C{0.767} & \C{0.768} & \C{0.757} & \C{0.721} & \C{0.793} \\
Gemma2-9B & 9B & \C{0.782} & \C{0.782} & \C{0.801} & \C{0.790} & \C{0.787} & \C{0.776} & \C{0.781} & \C{0.779} & \C{0.769} & \C{0.708} & \C{0.784} \\
\midrule
gte-Qwen2 & 7B & \C{0.597} & \C{0.605} & \C{0.604} & \C{0.609} & \C{0.579} & \C{0.588} & \C{0.602} & \C{0.620} & \C{0.605} & \C{0.605} & \C{0.611} \\
sfr-Mistral & 7B & \C{0.782} & \C{0.780} & \C{0.810} & \C{0.790} & \C{0.782} & \C{0.768} & \C{0.786} & \C{0.786} & \C{0.777} & \C{0.738} & \C{0.794} \\
Mistral-7B\textsubscript{Instruct} & 7B & \C{0.777} & \C{0.776} & \C{0.799} & \C{0.781} & \C{0.781} & \C{0.765} & \C{0.782} & \C{0.771} & \C{0.758} & \C{0.720} & \C{0.782} \\
bge-Gemma2 & 9B & \C{0.786} & \C{0.782} & \C{0.808} & \C{0.790} & \C{0.790} & \C{0.774} & \C{0.786} & \C{0.773} & \C{0.781} & \C{0.750} & \C{0.792} \\
\bottomrule
\end{tabular}
\caption{Original VQAScore for models using embeddings extracted from the last layer. We use text encoders from \href{https://huggingface.co/laion/CLIP-ViT-H-14-laion2B-s32B-b79K}{CLIP-ViT-H/14} (354M) and \href{https://huggingface.co/google-t5/t5-11b}{T5-XXL} (4.7B), along with four popular open-source pre-trained LLMs: \href{https://huggingface.co/Qwen/Qwen2-7B}{Qwen2} (7B), \href{https://huggingface.co/mistralai/Mistral-7B-v0.1}{Mistral-7B} (7B), \href{https://huggingface.co/meta-llama/Meta-Llama-3-8B}{Llama3} (8B), and \href{https://huggingface.co/google/gemma-2-9b}{Gemma2} (9B). Additionally, we include three embedding models fine-tuned on these LLMs: gte-Qwen2 (\href{https://huggingface.co/Alibaba-NLP/gte-Qwen2-7B-instruct}{gte-Qwen2-7B-instruct}; 7B), sfr-Mistral (\href{https://huggingface.co/Salesforce/SFR-Embedding-2_R}{SFR-Embedding-2\_R}; 7B), and bge-Gemma2 (\href{https://huggingface.co/BAAI/bge-multilingual-gemma2}{bge-multilingual-gemma2}; 9B). We also include an instruction fine-tuned model, \href{https://huggingface.co/mistralai/Mistral-7B-Instruct-v0.2}{Mistral-7B-Instruct} (7B).}
\label{tab:baseline_og}
\end{table*}
\setlength{\tabcolsep}{\oldtabcolsep}

\setlength{\tabcolsep}{3.5pt}
\begin{table*}[t]
\centering
\begin{tabular}{l@{\hspace{3mm}}lcccccccccccc}
\toprule
\textbf{Model} & \textbf{Layer} & \textbf{Avg} & \textbf{Attr.} & \textbf{Scene} & \textbf{Spat.} & \textbf{Action} & \textbf{Part} & \textbf{Count.} & \textbf{Comp.} & \textbf{Differ.} & \textbf{Neg.} & \textbf{Uni.}\\
\midrule
T5-XXL & 25 (last) & \C{0.795} & \C{0.795} & \C{0.816} & \C{0.803} & \C{0.803} & \C{0.780} & \C{0.799} & \C{0.800} & \C{0.793} & \C{0.739} & \C{0.804} \\
Mistral-7B & 33 (last) & \C{0.782} & \C{0.780} & \C{0.810} & \C{0.790} & \C{0.782} & \C{0.768} & \C{0.786} & \C{0.786} & \C{0.777} & \C{0.738} & \C{0.794} \\
Mistral-7B & 32 & \C{0.783} & \C{0.782} & \C{0.802} & \C{0.785} & \C{0.790} & \C{0.775} & \C{0.783} & \C{0.766} & \C{0.779} & \C{0.726} & \C{0.786} \\
Mistral-7B & 15 & \C{0.783} & \C{0.785} & \C{0.811} & \C{0.787} & \C{0.788} & \C{0.774} & \C{0.795} & \C{0.783} & \C{0.779} & \C{0.724} & \C{0.783} \\
Mistral-7B & 0\hphantom{0} (first) & \C{0.660} & \C{0.661} & \C{0.702} & \C{0.657} & \C{0.646} & \C{0.630} & \C{0.655} & \C{0.680} & \C{0.640} & \C{0.605} & \C{0.692} \\
\bottomrule
\end{tabular}
\caption{Original VQAScore for models using embeddings extracted from individual layers of \href{https://huggingface.co/mistralai/Mistral-7B-v0.1}{Mistral-7B} (7B). We include the baseline \href{https://huggingface.co/google-t5/t5-11b}{T5-XXL} (4.7B) model using embeddings extracted from the last layer as a reference.}
\label{tab:individualLayers_og}
\end{table*}
\setlength{\tabcolsep}{\oldtabcolsep}

\section{Training details}
\label{sec:training_details}

We follow the U-Net \cite{ronneberger_u-net_2015} based latent diffusion architecture from Stable Diffusion v2 \cite{Rombach_2022_CVPR} with a \href{https://github.com/mosaicml/diffusion}{replication training framework} by MosaicML \cite{mosaicml2023diffusion}. We use Diffusers as our main model codebase for the Variational Autoencoder (VAE), U-Net, and noise scheduler~\cite{von-platen-etal-2022-diffusers}. The configurations all follow the original Stable Diffusion~\cite{Rombach_2022_CVPR}. For each model, we swap the text encoder and freeze all components except for the U-Net. Before the text embeddings are input to the U-Net's cross-attention blocks, we apply a linear projection from each text encoder's embedding dimension to 1024 output features for all models. The base U-Net has $865{,}910{,}724$ trainable parameters, and the additional parameters from the projection for each model are shown in Table \ref{tab:emb_dims}.

\setlength{\tabcolsep}{3.8pt}
\begin{table*}[t]
\centering
\small
\begin{tabular}{lccccccccccccc}
\toprule
\textbf{Model} & \textbf{Embeddings} & \textbf{Avg} & \textbf{Attr.} & \textbf{Scene} & \textbf{Spat.} & \textbf{Action} & \textbf{Part} & \textbf{Count.} & \textbf{Comp.} & \textbf{Differ.} & \textbf{Neg.} & \textbf{Uni.}\\
\midrule
T5-XXL & last layer & \C{0.795} & \C{0.795} & \C{0.816} & \C{0.803} & \C{0.803} & \C{0.780} & \C{0.799} & \C{0.800} & \C{0.793} & \C{0.739} & \C{0.804} \\
T5-XXL & norm avg & \C{0.791} & \C{0.795} & \C{0.810} & \C{0.799} & \C{0.796} & \C{0.771} & \C{0.795} & \C{0.813} & \C{0.783} & \C{0.730} & \C{0.799} \\
\midrule
bge-Gemma2 & last layer & \C{0.786} & \C{0.782} & \C{0.808} & \C{0.790} & \C{0.790} & \C{0.774} & \C{0.786} & \C{0.773} & \C{0.781} & \C{0.750} & \C{0.792} \\
bge-Gemma2 & avg & \C{0.809} & \C{0.807} & \C{0.822} & \C{0.815} & \C{0.814} & \C{0.793} & \C{0.811} & \C{0.813} & \C{0.794} & \C{0.755} & \C{0.807} \\
bge-Gemma2 & norm avg & \C{0.801} & \C{0.801} & \C{0.821} & \C{0.806} & \C{0.806} & \C{0.790} & \C{0.803} & \C{0.805} & \C{0.789} & \C{0.758} & \C{0.813} \\
\midrule
Mistral-7B & last layer & \C{0.782} & \C{0.780} & \C{0.810} & \C{0.790} & \C{0.782} & \C{0.768} & \C{0.786} & \C{0.786} & \C{0.777} & \C{0.738} & \C{0.794} \\
Mistral-7B & avg & \C{0.786} & \C{0.785} & \C{0.809} & \C{0.797} & \C{0.793} & \C{0.774} & \C{0.783} & \C{0.780} & \C{0.761} & \C{0.726} & \C{0.788} \\
Mistral-7B & norm avg & \C{0.799} & \C{0.798} & \C{0.820} & \C{0.808} & \C{0.810} & \C{0.789} & \C{0.791} & \C{0.796} & \C{0.790} & \C{0.734} & \C{0.805} \\
\bottomrule
\end{tabular}
\caption{Original VQAScore for models using different embedding strategies: standard last-layer embeddings (\textit{last layer}), average embeddings across all layers (\textit{avg}), and average embeddings across all normalized layers (\textit{norm avg}). We evaluate the encoder from \href{https://huggingface.co/google-t5/t5-11b}{T5-XXL} (4.7B), the pre-trained LLM \href{https://huggingface.co/mistralai/Mistral-7B-v0.1}{Mistral-7B} (7B), and the fine-tuned embedding model bge-Gemma2 (\href{https://huggingface.co/BAAI/bge-multilingual-gemma2}{bge-multilingual-gemma2}; 9B).}
\label{tab:averaging_og}
\end{table*}
\setlength{\tabcolsep}{\oldtabcolsep}

\setlength{\tabcolsep}{4.5pt}
\begin{table*}[t!]
\centering
\begin{tabular}{lcccccccccccc}
\toprule
\textbf{Model} & \textbf{Avg} & \textbf{Attr.} & \textbf{Scene} & \textbf{Spat.} & \textbf{Action} & \textbf{Part} & \textbf{Count.} & \textbf{Comp.} & \textbf{Differ.} & \textbf{Neg.} & \textbf{Uni.}\\
\midrule
bge-Gemma2 & \C{0.786} & \C{0.782} & \C{0.808} & \C{0.790} & \C{0.790} & \C{0.774} & \C{0.786} & \C{0.773} & \C{0.781} & \C{0.750} & \C{0.792} \\
bge-Gemma2\textsubscript{pooled} & \C{0.802} & \C{0.801} & \C{0.828} & \C{0.807} & \C{0.806} & \C{0.789} & \C{0.812} & \C{0.806} & \C{0.799} & \C{0.763} & \C{0.835} \\
\midrule
sfr-Mistral & \C{0.782} & \C{0.780} & \C{0.810} & \C{0.790} & \C{0.782} & \C{0.768} & \C{0.786} & \C{0.786} & \C{0.777} & \C{0.738} & \C{0.794} \\
sfr-Mistral\textsubscript{pooled} & \C{0.782} & \C{0.778} & \C{0.810} & \C{0.779} & \C{0.788} & \C{0.774} & \C{0.778} & \C{0.772} & \C{0.772} & \C{0.739} & \C{0.796} \\
\bottomrule
\end{tabular}
\caption{Original VQAScore for models using embeddings extracted from the last layer compared to models with additional conditioning on global pooled embeddings \cite{podell_sdxl_2023}. We evaluate the fine-tuned embedding models bge (\href{https://huggingface.co/BAAI/bge-multilingual-gemma2}{bge-multilingual-gemma2}; 9B) and sfr (\href{https://huggingface.co/Salesforce/SFR-Embedding-2_R}{SFR-Embedding-2\_R}; 7B).}
\label{tab:pooling_og}
\end{table*}
\setlength{\tabcolsep}{\oldtabcolsep}

\setlength{\tabcolsep}{4.8pt}
\begin{table*}[t!]
\centering
\begin{tabular}{lccccccccccccc}
\toprule
\textbf{Model} & \textbf{Size} & \textbf{Avg} & \textbf{Attr.} & \textbf{Scene} & \textbf{Spat.} & \textbf{Action} & \textbf{Part} & \textbf{Count.} & \textbf{Comp.} & \textbf{Differ.} & \textbf{Neg.} & \textbf{Uni.}\\
\midrule
Qwen2 & 1.5B & \C{0.758} & \C{0.759} & \C{0.784} & \C{0.761} & \C{0.760} & \C{0.738} & \C{0.756} & \C{0.753} & \C{0.747} & \C{0.704} & \C{0.755} \\
Qwen2 & 7B & \C{0.772} & \C{0.772} & \C{0.798} & \C{0.777} & \C{0.782} & \C{0.761} & \C{0.760} & \C{0.759} & \C{0.748} & \C{0.712} & \C{0.785} \\
\midrule
Gemma2 & 2B & \C{0.770} & \C{0.773} & \C{0.797} & \C{0.774} & \C{0.774} & \C{0.759} & \C{0.783} & \C{0.764} & \C{0.756} & \C{0.710} & \C{0.787} \\
Gemma2 & 9B & \C{0.782} & \C{0.782} & \C{0.801} & \C{0.790} & \C{0.787} & \C{0.776} & \C{0.781} & \C{0.779} & \C{0.769} & \C{0.708} & \C{0.784} \\
\bottomrule
\end{tabular}
\caption{Original VQAScore for models with different LLM sizes, using embeddings extracted from the last layer. We evaluate the pre-trained LLMs: \href{https://huggingface.co/Qwen/Qwen2-1.5B}{Qwen2} (1.5B), \href{https://huggingface.co/Qwen/Qwen2-7B}{Qwen2} (7B), \href{https://huggingface.co/google/gemma-2-2b}{Gemma2} (2B), \href{https://huggingface.co/google/gemma-2-9b}{Gemma2} (9B).}
\label{tab:modelSizes_og}
\end{table*}
\setlength{\tabcolsep}{\oldtabcolsep}

For training, we use a 46 million text-image pair subset of the LAION-Aesthetics dataset \cite{schuhmann_laion-5b_2022}. We perform center cropping on all training images and LLM-based caption upsampling with VisualFactChecker (VFC) \cite{ge_visual_2024}. We train all models for $800{,}000$ iterations at $256{\times}256$ resolution with a global batch size of $2{,}048$ on $32$ A100 GPUs. We use a caption drop probability of $0.1$ and use the AdamW optimizer \cite{loshchilov2019decoupledweightdecayregularization} with a learning rate of $10^{-4}$ and weight decay of $0.01$. We also use additional optimizations such as FlashAttention \cite{dao2022flashattentionfastmemoryefficientexact}, \href{https://docs.mosaicml.com/projects/composer/en/stable/method_cards/low_precision_groupnorm.html}{half precision GroupNorm}, \href{https://docs.mosaicml.com/projects/composer/en/stable/method_cards/low_precision_layernorm.html}{half precision LayerNorm}, and Fully Sharded Data Parallel's (FSDP)~\cite{zhao2023pytorchfsdpexperiencesscaling} \textit{SHARD\_GRAD\_OP} mode for enhanced GPU scaling~\cite{mosaicml2023diffusion}. Notably, we also pre-compute the VAE image latents and text embeddings for the entire training dataset to reduce repetitive compute time and costs, with each model taking around $7$ days to fully train.

\section{Prompt upsampling}
\label{sec:upsampling}

\subsection{Prompting}

\lstset{
basicstyle=\small\ttfamily,
columns=flexible,
breaklines=true
}

We use the instruction fine-tuned \href{https://huggingface.co/google/gemma-2-9b-it}{Gemma2-9B} \cite{team2024gemma} to perform prompt upsampling at inference. We use the following prompt:
\begin{lstlisting}
You are part of a team that creates images from texts. You work with an assistant that will draw anything you say. For example, outputting "a beautiful morning in the woods with the sun peaking through the trees" will trigger your partner to output an image of a forest morning, as you described.

You are responsible for helping people create detailed, amazing images. The way to accomplish this is to take their short prompts and make them extremely detailed and descriptive.

There are a few rules to follow:
- Try to start with the original input prompt in your output, and then add details after.
- You will only ever output a single image description per user request.
- Image descriptions must be between 15-80 words. Extra words will be ignored.
\end{lstlisting}

With the following generation parameters: 

\begin{lstlisting}
do_sample=True
temperature=0.01
top_p=1.0
max_new_tokens=1024
\end{lstlisting}

\subsubsection{Examples}
In this subsection, we list a few examples of the upsampled prompts from GenAI-Bench \cite{lin_evaluating_2024} that are used in Figure \ref{fig:heatmap}:

\textit{A tomato vine with several tomatoes on it, all yellow except the largest which is red} $\Rightarrow$ ``A tomato vine with several tomatoes on it, all yellow except the largest which is red. The vine is thick and green, with leaves that are a deep, rich green. The tomatoes are plump and juicy, glistening in the sunlight. The largest tomato, a vibrant red, hangs heavy on the vine, contrasting beautifully with the yellow fruit around it.''

\textit{A tomato vine with several tomatoes on it, all yellow except the largest which is red} $\Rightarrow$ ``A larger gorilla hands a smaller mechanical monkey a banana. The gorilla is silverback, with thick fur and a wise expression in its eyes. The mechanical monkey is made of polished brass, with intricate gears visible on its chest and limbs. It has glowing red eyes and a mischievous grin. The banana is ripe and yellow, held out in the gorilla's massive hand. The background is a lush jungle, with vines and ferns creating a vibrant tapestry.''

\section{VQAScore evaluation}
\label{sec:variation}

\subsection{VQAScore details}
VQAScore is a metric for evaluating how well a generated image semantically aligns with its text prompt by using visual-question-answering (VQA). It ranges from $0$ to $1$, where scores closer to $1$ represent close alignment with the prompt, and 0 means the generated image doesn't at all. Intuitively, it measures how well the image understands and represents the prompt, going beyond surface-level similarity. Please refer to the original VQAScore paper \cite{lin_evaluating_2024} for further insight.

\subsection{Additional layer-normalized averaging results}
We report additional results for our models using layer-normalized average embeddings, which aggregate representations across all layers. Table \ref{tab:norm_avg_models} presents a comprehensive comparison with the baseline T5 model and models utilizing last-layer embeddings.

\subsection{Original CLIP-FlanT5 model}
We show results for our models evaluated using the original VQAScore implementation in Tables \ref{tab:baseline_og}, \ref{tab:individualLayers_og}, \ref{tab:averaging_og}, \ref{tab:pooling_og}, \ref{tab:modelSizes_og}. As can be seen in these tables, the \href{https://github.com/linzhiqiu/CLIP-FlanT5}{custom CLIP-FlanT5 model} introduced in VQAScore paper \cite{lin_evaluating_2024} is not as capable as GPT-4o \cite{openai_gpt-4o_2024} in discriminating between different models, but still show correlated trends to the GPT-4o results.

\subsection{Our GPT-4o implementation}
We build upon VQAScore's support for GPT-4v by replicating the \href{https://github.com/linzhiqiu/t2v_metrics/blob/main/t2v_metrics/models/vqascore_models/gpt4v_model.py}{code} and \href{https://github.com/linzhiqiu/t2v_metrics/blob/main/gpt4_eval.py}{swapping} in the GPT-4o API instead. We also limit the number of tokens returned by setting \textit{top\_logprobs} $=20$. We found that a simple retry up to $3$ times eliminated almost all errors, such as timeout and invalid answer token selections. Apart from enhanced performance, GPT-4o also includes support for prompt caching, allowing for reduced time and costs.

\subsection{Random variation}
When computing VQAScore with GPT-4o, we generate the $1600$ images of upsampled GenAI-Bench prompts for each model. The variation resulting from the choice of random seeds is in the order of $\pm0.004$, depending on the category. The variation from the same seed is $\pm0.003$, from non-deterministic CUDA operations and possible non-determinism of the GPT-4o queries.

\section{More visual comparisons}
\label{sec:visuals}
\begin{figure*}[p]
    \centering
    \begin{tabular}{cccc}
        {\centering CLIP (\emph{last layer})} & {\centering T5-XXL (\emph{last layer})} & {\centering Mistral (\emph{norm avg})} & {\centering bge-Gemma2 (\emph{norm avg})} \\[2mm] 
        \includegraphics[width=0.245\textwidth]{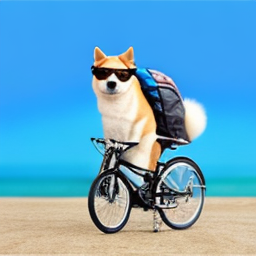} &
        \includegraphics[width=0.245\textwidth]{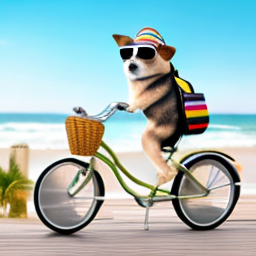} &
        \includegraphics[width=0.245\textwidth]{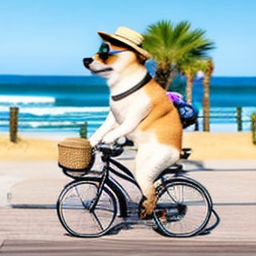} &
        \includegraphics[width=0.245\textwidth]{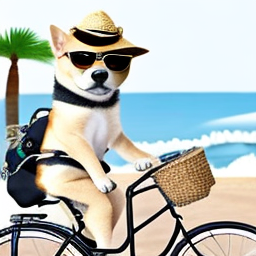} \\[1mm]
        \multicolumn{4}{c}{\emph{A photo of a Shiba Inu dog with a backpack riding a bike. It is wearing sunglasses and a beach hat.}} \\[3mm] 
        \includegraphics[width=0.245\textwidth]{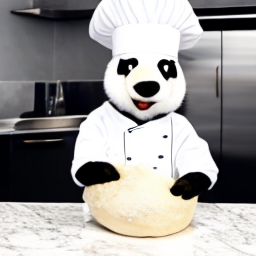} &
        \includegraphics[width=0.245\textwidth]{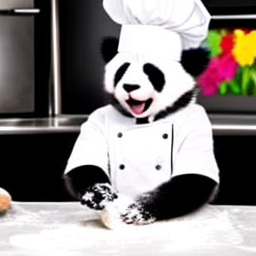} &
        \includegraphics[width=0.245\textwidth]{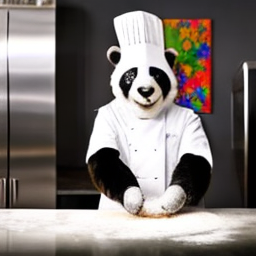} &
        \includegraphics[width=0.245\textwidth]{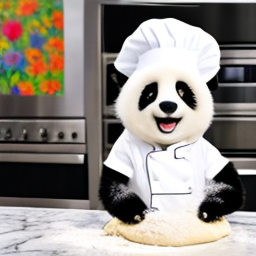} \\[1mm]
        \multicolumn{4}{c}{\emph{A high contrast portrait of a very happy fuzzy panda dressed as a chef in a high end kitchen making dough.}} \\[1.5mm]
        \multicolumn{4}{c}{\emph{ There is a painting of flowers on the wall behind him.}} \\[3mm]

        \includegraphics[width=0.245\textwidth]{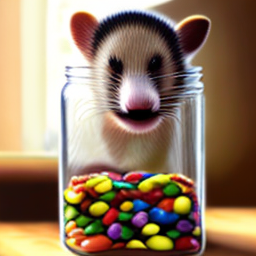} &
        \includegraphics[width=0.245\textwidth]{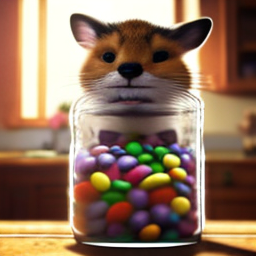} &
        \includegraphics[width=0.245\textwidth]{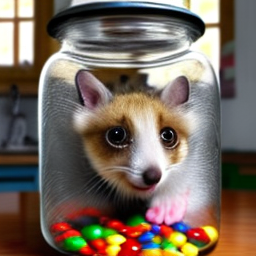} &
        \includegraphics[width=0.245\textwidth]{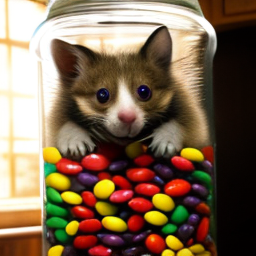} \\[1mm]
        \multicolumn{4}{c}{\emph{A mischievous ferret with a playful grin squeezes itself into a large glass jar, surrounded by colorful candy.}} \\[1.5mm] 
        \multicolumn{4}{c}{\emph{The jar sits on a wooden table in a cozy kitchen, and warm sunlight filters through a nearby window.}} \\[3mm] 
        \includegraphics[width=0.245\textwidth]{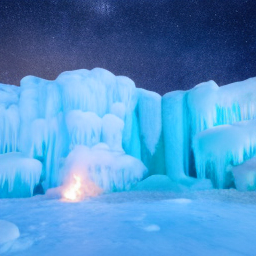} &
        \includegraphics[width=0.245\textwidth]{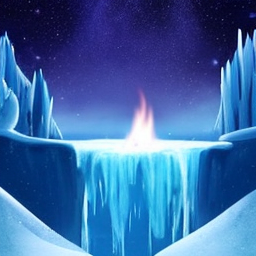} &
        \includegraphics[width=0.245\textwidth]{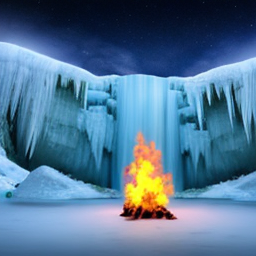} &
        \includegraphics[width=0.245\textwidth]{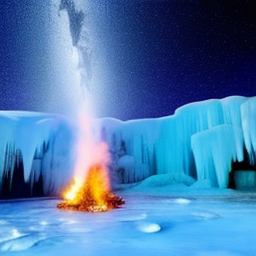} \\[1mm]
        \multicolumn{4}{c}{\emph{An icy landscape under a starlit sky, where a magnificent frozen waterfall flows over a cliff. In the center of the scene,}} \\[1.5mm]
        \multicolumn{4}{c}{\emph{a fire burns bright, its flames seemingly frozen in place, casting a shimmering glow on the surrounding ice and snow.}} 
    
\end{tabular}\vspace{-1mm}%
\caption{Visual comparison of images generated with different text encoders. We use last-layer embeddings (\textit{last layer}) from the text encoders of \href{https://huggingface.co/laion/CLIP-ViT-H-14-laion2B-s32B-b79K}{CLIP-ViT-H/14} (354M) and \href{https://huggingface.co/google-t5/t5-11b}{T5-XXL} (4.7B). We also use average layer-normalized embeddings (\textit{norm avg}) from the pre-trained LLM \href{https://huggingface.co/mistralai/Mistral-7B-v0.1}{Mistral-7B} (7B) and the fine-tuned embedding model bge-Gemma2 (\href{https://huggingface.co/BAAI/bge-multilingual-gemma2}{bge-multilingual-gemma2}; 9B).}
\label{fig:visual_comparison_popular}
\end{figure*}

\begin{figure*}[p]
    \centering
    \begin{tabular}{cccc}
        {\centering CLIP (\emph{last layer})} & {\centering T5-XXL (\emph{last layer})} & {\centering Mistral (\emph{norm avg})} & {\centering bge-Gemma2 (\emph{norm avg})} \\[2mm] 
        \includegraphics[width=0.245\textwidth]{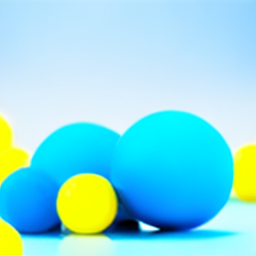} &
        \includegraphics[width=0.245\textwidth]{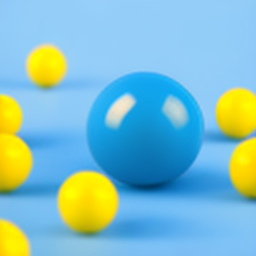} &
        \includegraphics[width=0.245\textwidth]{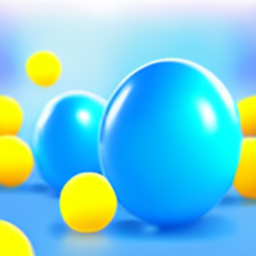} &
        \includegraphics[width=0.245\textwidth]{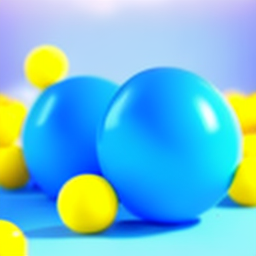} \\[1mm]
        \multicolumn{4}{c}{\emph{A scene with two blue balls amidst many yellow ones. The blue balls are slightly larger than}} \\[1.5mm] 
        \multicolumn{4}{c}{\emph{the yellow ones and have a smooth, glossy surface that reflects the light.}} \\[2.5mm] 
        \includegraphics[width=0.245\textwidth]{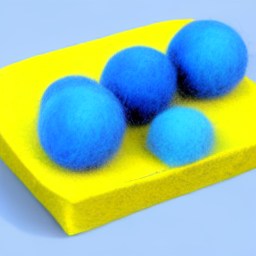} &
        \includegraphics[width=0.245\textwidth]{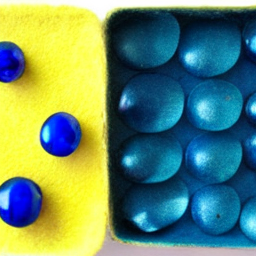} &
        \includegraphics[width=0.245\textwidth]{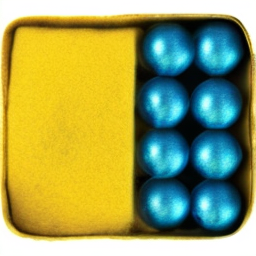} &
        \includegraphics[width=0.245\textwidth]{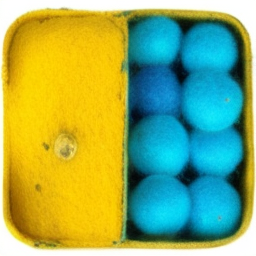} \\[1mm]
        \multicolumn{4}{c}{\emph{A yellow felt box has no metallic blue spheres on the left side and has blue metallic spheres on the right side.}} \\[3mm]

        \includegraphics[width=0.245\textwidth]{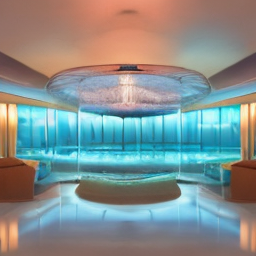} &
        \includegraphics[width=0.245\textwidth]{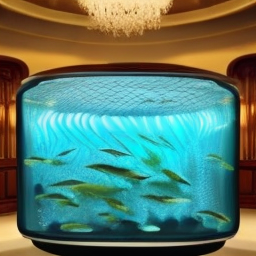} &
        \includegraphics[width=0.245\textwidth]{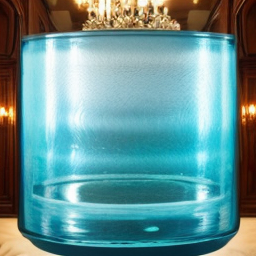} &
        \includegraphics[width=0.245\textwidth]{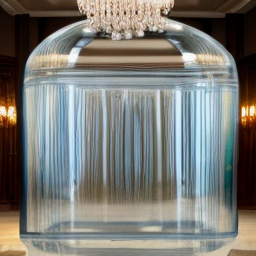} \\[1mm]
        \multicolumn{4}{c}{\emph{There is a large fish aquarium in the center of the luxurious living room, but there are no fish in it.}} \\[1.5mm] 
        \multicolumn{4}{c}{\emph{The aquarium is made of polished, rippling glass, reflecting the warm glow of the chandelier above.}} \\[3mm] 
        \includegraphics[width=0.245\textwidth]{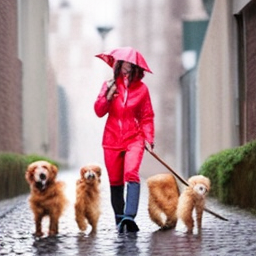} &
        \includegraphics[width=0.245\textwidth]{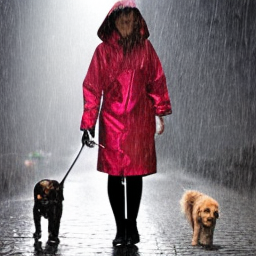} &
        \includegraphics[width=0.245\textwidth]{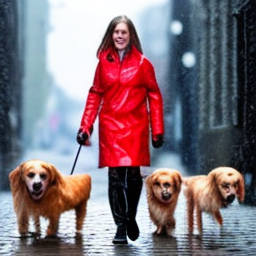} &
        \includegraphics[width=0.245\textwidth]{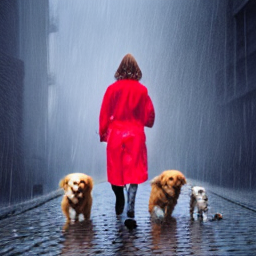} \\[1mm]
        \multicolumn{4}{c}{\emph{A woman with three dogs and no umbrella in the drizzle. Two golden retrievers bound ahead,}} \\[1.5mm]
        \multicolumn{4}{c}{\emph{their tails wagging despite the light rain, while a small terrier trots obediently by her side.}} 
    
\end{tabular}\vspace{-1mm}%
\caption{Visual comparison of images generated with different text encoders. We use last-layer embeddings (\textit{last layer}) from the text encoders of \href{https://huggingface.co/laion/CLIP-ViT-H-14-laion2B-s32B-b79K}{CLIP-ViT-H/14} (354M) and \href{https://huggingface.co/google-t5/t5-11b}{T5-XXL} (4.7B). We also use average layer-normalized embeddings (\textit{norm avg}) from the pre-trained LLM \href{https://huggingface.co/mistralai/Mistral-7B-v0.1}{Mistral-7B} (7B) and the fine-tuned embedding model bge-Gemma2 (\href{https://huggingface.co/BAAI/bge-multilingual-gemma2}{bge-multilingual-gemma2}; 9B).}
\label{fig:additional_prompts}
\end{figure*}

We provide additional visual comparisons between the baseline CLIP and T5 models using last-layer embeddings with the Mistral and bge-Gemma2 models using layer-normalized average embeddings. Figure \ref{fig:visual_comparison_popular} shows examples from common text-to-image prompts, and Figure \ref{fig:additional_prompts} shows additional prompts from GenAI-Bench.

\setlength{\tabcolsep}{1pt}
\renewcommand{\arraystretch}{0}
\setlength{\tabcolsep}{\oldtabcolsep}
\renewcommand{\arraystretch}{1}

\setlength{\tabcolsep}{1pt}
\renewcommand{\arraystretch}{0}
\setlength{\tabcolsep}{\oldtabcolsep}
\renewcommand{\arraystretch}{1}

\end{document}